\documentclass[lettersize,journal]{IEEEtran}

\usepackage{amsmath,amsfonts}
\usepackage{algorithm}
\usepackage{array}
\usepackage{textcomp}
\usepackage{stfloats}
\usepackage{url}
\usepackage{verbatim}
\usepackage{subcaption}
\usepackage{graphicx}
\usepackage{cite}
\usepackage{tikz}
\usetikzlibrary{shapes}
\usepackage[export]{adjustbox}
\usepackage{contour}
\usepackage[normalem]{ulem}
\usepackage{pifont}
\usepackage{xcolor}

\usepackage{graphicx}
\usepackage{tabularx}
\usepackage{booktabs}
\usepackage{multirow}
\usepackage{amssymb}
\usepackage{mathtools}
\usepackage{makecell}
\usepackage{caption}
\usepackage{algpseudocode}
\renewcommand{\vec}[1]{\boldsymbol{#1}}
\newcommand{\transp}{^\top}
\renewcommand{\matrix}[1]{#1}

\DeclareRobustCommand\sampleline[1]{%
  \tikz\draw[#1] (0,0) (0,\the\dimexpr\fontdimen22\textfont2)
  -- (1.5em,\the\dimexpr\fontdimen22\textfont2);%
}

\newcommand{\splitline}[4]{
\tikz{\draw[#1, #3, #4] (0, 0) (0,  \the\dimexpr\fontdimen22\textfont2) -- (0.75em,  \the\dimexpr\fontdimen22\textfont2) edge[#2] +(0:0.75em);}
}

\definecolor{green}{RGB}{0, 170, 0}
\definecolor{red}{RGB}{200, 0, 0}
\definecolor{lightgray}{RGB}{200, 200, 200}
\definecolor{darkgray}{RGB}{150, 150, 150}
\definecolor{lightblue}{RGB}{41, 200, 200}
\definecolor{darkblue}{RGB}{41, 92, 207}
\definecolor{jump_red}{RGB}{176, 33, 33}
\definecolor{jump_yellow}{RGB}{250, 191, 5}
\definecolor{steelblue}{RGB}{70, 130, 180}
\definecolor{firebrick}{RGB}{178, 34, 34}

\newcommand{\redtriangle}{\raisebox{0.5pt}{\tikz{\node[draw=none,scale=0.5,regular polygon, regular polygon sides=3,fill=firebrick,rotate=0](){};}}}
\newcommand{\bluetriangle}{\raisebox{0.5pt}{\tikz{\node[draw=none,scale=0.5,regular polygon, regular polygon sides=3,fill=steelblue,rotate=0](){};}}}
\def\checkmark{\tikz\fill[scale=0.4](0,.35) -- (.25,0) -- (1,.7) -- (.25,.15) -- cycle;}

\contourlength{0.8pt}

\usepackage{array}
\newcolumntype{L}[1]{>{\raggedright\let\newline\\\arraybackslash\hspace{0pt}}p{#1}}
\newcolumntype{C}[1]{>{\centering\let\newline\\\arraybackslash\hspace{0pt}}p{#1}}
\newcolumntype{R}[1]{>{\raggedleft\let\newline\\\arraybackslash\hspace{0pt}}p{#1}}

\begin{document}

\title{ASAP-MPC: An Asynchronous Update Scheme \\ for Online Motion Planning with \\ Nonlinear Model Predictive Control}
% for Nonlinear Model Predictive Control in Motion Planning}

\author{\IEEEauthorblockN{Dries Dirckx\textsuperscript{1}, Mathias Bos\textsuperscript{1}, Bastiaan Vandewal\textsuperscript{1}, Lander Vanroye, Wilm Decr\'{e}, Jan Swevers} \\
        % <-this % stops a space
%\thanks{Flanders Maaaaaake - ARENA, FROGS, DIRAC, FLEXMOSYS}% <-this % stops a space
\IEEEauthorblockA{\textit{MECO Research Team, Department of Mechanical Engineering, KU Leuven, Leuven, Belgium} \\
\textit{Flanders Make@KU Leuven, Leuven, Belgium}\\
Email: firstname.lastname@kuleuven.be}
\thanks{\textsuperscript{1}These authors contributed equally.}
}
% \address{\textit{MECO Research Team, Department of Mechanical Engineering,}\\
% \textit{KU Leuven}, Belgium (e-mail: firstname.lastname@kuleuven.be)\\
% and \textit{Flanders Make@KU Leuven}, Belgium}



\maketitle

\begin{abstract}

This paper presents a Nonlinear Model Predictive Control (NMPC) scheme targeted at motion planning for mechatronic motion systems, such as drones and mobile platforms. NMPC-based motion planning typically requires low computation times to be able to provide control inputs at the required rate for system stability, disturbance rejection, and overall performance. Although there exist various ways in literature to reduce the solution times in NMPC, such times may not be low enough to allow real-time implementations. This paper presents ASAP-MPC, an approach to handle varying, sometimes restrictively large, solution times with an asynchronous update scheme, always allowing for full convergence and real-time execution. The NMPC algorithm is combined with a linear state feedback controller tracking the optimised trajectories for improved robustness against possible disturbances and plant-model mismatch. ASAP-MPC seamlessly merges trajectories, resulting from subsequent NMPC solutions, providing a smooth and continuous overall trajectory for the motion system. This framework's applicability to embedded applications is shown on two different experiment setups where a state-of-the-art method fails: a quadcopter flying through a cluttered environment in hardware-in-the-loop simulation and a scale model truck-trailer manoeuvring in a structured lab environment.
\end{abstract}

\def\abstractname{Note to Practitioners}
\begin{abstract}
This letter is motivated by the need for a robust trajectory planning scheme for embedded autonomous motion planning. Various methods rapidly compute trajectories but are not sufficiently robust for complex applications or do not guarantee stability of the mechatronic system. This letter introduces ASAP-MPC, an innovative Nonlinear Model Predictive Control (NMPC) scheme integrating NMPC with a stabilising, linear state feedback controller.
This scheme facilitates the combination of optimal control for motion planning and well-known classical feedback control on embedded platforms, for more complex systems, lower computational power requirements, and faster prototyping.
Target applications are mobile ground and aerial robots, and robot manipulators, in contexts where there is a requirement for online re-planning and accurate tracking of the planned trajectory.
\end{abstract}

\begin{IEEEkeywords}
Motion Control, Integrated Planning and Control, Constrained Motion Planning, Optimization and Optimal Control
% Trajectory and Path Planning, Trajectory Tracking and Path Following, Optimal Motion Planning and Control, Model Predictive Control
\end{IEEEkeywords}

% ===========================================
% CONTENT

% !TeX root = main.tex
\section{Introduction}
\label{section:introduction}

\IEEEPARstart{D}{espite} a lot of research effort in the past few decades, online motion (re-)planning is still a challenge for complex motion systems to autonomously perform difficult tasks. Autonomous path and trajectory planning is one of the cornerstones in robotics to safely deploy aerial robots, ground robots, and robot manipulators. The kinematics and dynamics of these systems, and a possibly changing environment require trajectory planning, i.e. simultaneously planning the geometric path and including timing information along this path, rather than simply planning collision-free geometrical paths through static environments. However, such kinematics and dynamics, the various objectives and the often non-convex constraints imposed by the use cases, render the planning problem challenging to solve, specifically in an online re-planning context with limited computational power available.\\

% Path planning and trajectory planning are terms widely used in motion planning research. Path planning covers the problem of finding a geometrical path through an environment by connecting discrete Cartesian points, without any associated timing information. On the contrary, trajectory planning also includes the time instances at which these points have to be reached and the control inputs to move the system between those points. \\

The most popular techniques to tackle path or trajectory planning are graph-search methods, (deep) reinforcement learning (RL), and optimisation-based planning. Graph-search algorithms, such as~\cite{rrt_star} and~\cite{a_star}, have the advantages of asymptotic completeness and a derivative-free evaluation. They often suffer from long evaluation times, and therefore typically are not used for online re-planning but rather for computing a path offline and tracking it afterwards. The dynamics are not directly considered in the path generation, and therefore an extra smoothing step is generally required. RL has gained a lot of popularity as its learned policy is fast to evaluate online, it can be used in a model-free way, and can naturally handle highly complex dynamics and tasks. Its downsides are the requirement of large training data sets collected through interaction with the environment, and the absence of a feasibility guarantee for the obtained control inputs. Furthermore, the behaviour is often unpredictable for system states that were not seen during training. A survey of learning-based architectures for motion planning can be found in~\cite{aradi}.\\

Optimisation-based planning, based on an Optimal Control Problem (OCP) formulation, as presented in~\cite{ocp_kirk}, can be deployed online with receding horizon re-planning techniques, i.e. Model Predictive Control (MPC), as described in~\cite{mpc_camacho} and~\cite{rawlings}. The OCP formulation directly includes the robot kinematics or dynamics model, and outputs feasible control inputs and/or trajectories. This technique is also applicable to unseen environments since there is no need for training data. The main drawback is the computational cost during the online evaluation, which is negatively impacted by the complexity of the dynamics and the task. More complex dynamic models can improve the performance, but often cause a significant computational delay larger than the required sampling time in an online MPC application. \\
% \red{Coping with computational delays is still an active challenge for complex trajectory planning tasks, multiple methods have been designed in an attempt to solve it.}

This work focuses on the class of systems with a tight requirement on the trajectory tracking accuracy, e.g. motion planning for obstacle avoidance, and for which stiff feedback control is possible. Realising collision-free motion within a complex, dynamic environment for this class of systems requires solving a combination of three sub-problems: (1)~a geometric problem of avoiding collision, (2)~generating dynamically feasible motions, and (3)~accurately tracking and executing the planned motion. In literature, MPC is presented as a strategy to tackle all three problems simultaneously. However, given its computational cost, it is a challenge to design the MPC such that it is able to solve all three sub-problems at a control sampling rate that is sufficiently fast in order to effectively compensate disturbances.

Considering these advantages, disadvantages, and challenges, it is clear that MPC has the potential to be a robust technique for complex motion planning problems, but only if a strategy is found to robustly cope with the computational delay, worsened by the problem complexity and limited computational power in embedded applications. \\

This paper presents a scheme that implements such a strategy, allowing extra time to find a solution while still guaranteeing a smooth and feasible overall trajectory. It makes the following contributions:
\begin{itemize}
    \item[\ding{43}] Introducing ASAP-MPC as a scheme that
        \begin{itemize}
            \item explicitly {\textbf{copes with computational delay}} in MPC,
            \item naturally integrates {\textbf{linear feedback}} at the required rate,
            \item smoothly connects subsequent trajectories through future {\textbf{on-trajectory updates}}, and
            \item{\textbf{rapidly incorporates new environmental information}} while allowing more time to find a solution to the OCP at critical points, through an asynchronous update scheme.
        \end{itemize}
    \item[\ding{43}] Showing its functionality through simulation and lab experiments on two significantly different applications: an \textbf{aerial robot} flying through a cluttered, dynamic environment with quickly moving obstacles, and an \textbf{autonomous mobile robot (AMR)} consisting of \textbf{a truck with a trailer} performing multiple parking manoeuvres.%These applications are both significantly different from a dynamic model point of view, and from an OCP formulation point of view. The latter is different in the sense that the aerial vehicle problem is a single-stage, fixed time problem with explicit obstacle avoidance constraints, whereas the truck-trailer AMR problem is a multi-stage, free time problem with stage-dependent box constraints on the position.
    \item[\ding{43}] Displaying its applicability to complex situations with nonlinear systems where a state-of-the-art method fails.
    \end{itemize}

The paper is structured as follows. First, Section~\ref{section:related_work} presents related work on different strategies to approach optimisation-based motion planning. Next, Section~\ref{section:ASAP} presents the ASAP-MPC methodology and compares it to state-of-the-art MPC variants. Section~\ref{section:experiment_setups} introduces the experiment setups to showcase the method's applicability. Finally, Section~\ref{section:results} discusses the results.%and Section~\ref{section:conclusion} lists the most important conclusion and some future improvements.
% !TeX root = main.tex
\section{Related Work}
\label{section:related_work}

Nonlinear Model Predictive Control (NMPC) periodically solves an OCP to find control sequences and state trajectories that have to (1)~stabilise a nonlinear motion system, (2)~agilely control it, (3)~reject disturbances acting on it, and (4)~safely react to environmental changes~\cite{rawlings}. Furthermore, NMPC assumes that the solution to the OCP is instantaneously available. However, any NMPC implementation always involves a (significant) finite computation time. In mechatronic applications, with typical control sampling rates in the range of 10 to 1000 Hz, guaranteeing a feasible, locally optimal solution at these control rates is a challenging task, especially for practical applications that involve nonlinear systems, long planning horizons, complex environments, or limited onboard computational power. Four possible solutions have emerged in literature to deal with this task:
\begin{center}
    \begin{itemize}
        \item constructing a faster/task-tailored solver,
        \item formulating an approximate problem that is easier to solve
        \item early-termination of the algorithm before convergence is reached,
        \item allowing for computational delay.
    \end{itemize}
\end{center}

\subsection{Fast Solvers and Specific Formulations}
A considerable amount of research effort focuses on finding faster and task-tailored solvers for NMPC problems. In~\cite{obayashi2018}, Obayashi and Takano implement a two-step approach of an approximate Nonlinear Programming (NLP) problem with an active-set Sequential Quadratic Programming (SQP) method to improve computational speed. The approximate problem finds an initial point close to feasibility for the original problem, decreasing the number of active inequality constraints. In a second step, an SQP-type method with an active set QP-solver is used, combined with that initial point, to quickly find a solution to the original problem. SQP methods greatly benefit from a proper initial guess close to a regular solution of the original problem~\cite{rawlings}. They report a decrease in computation time by a factor 30 compared to fmincon (MATLAB) and a smaller difference between the average and the worst computation time. 

GRAMPC~\cite{grampc} is an MPC solver based on an augmented Lagrangian formulation for dynamical systems in contexts requiring embedded MPC. It combines the augmented Lagrangian with an inner minimisation problem, that employs a gradient method, to find solutions within a (sub)millisecond time frame. The paper demonstrates the improved performance compared with well-known MPC frameworks such as ACADO~\cite{acado} and VIATOC~\cite{viatoc} on a set of benchmark applications that require MPC running on embedded hardware. For instance, for the quadrotor benchmark from~\cite{grampc_1st_paper}, GRAMPC computes the solution up to ten times faster than ACADO and VIATOC. For some of the problems, GRAMPC sacrifices accuracy for speed but it is argued that for MPC fast re-planning is more crucial than absolute accuracy.

A different approach is taken by the first-order, matrix-free PANOC solver~\cite{panoc}, which does not require Hessian information nor inner iterations to find a solution to an unconstrained nonlinear MPC problem. In~\cite{ajay2018}, Sathya et al. illustrate PANOC's capabilities on a lab-scale embedded system, driving a mobile robot collision-free through an environment with obstacles, with an MPC controller (using PANOC) running at 10~Hz. It reports a decrease in computation times of approximately two orders of magnitude compared to well-known NLP solvers IPOPT and fmincon (MATLAB).
%{\color{red} Find PANOC, David paper, GRAMPC, ACADOS, Alpaqa-NLP, HILO-MPC, Obayashi, }

\subsection{Premature Solver Interruptions}
The examples so far show that the emergence of fast NLP solver algorithms enables the use of MPC for applications of increasing complexity, by decreasing the computation times required to solve their related OCP. However, one might argue that faster solvers not only solve an OCP in a shorter time, but they could also solve an even more complex problem in the same time. For the latter situation, a way to cope with the significant computational delay must still be sought.

The Real-Time Iteration (RTI) scheme for NMPC~\cite{diehl_rti} abandons the idea of solving the OCP to convergence within one time step, but rather spreads out the convergence over multiple time steps, while using the intermediate non-converged solutions as control inputs. They argue that a receding horizon scheme entails solving very similar subsequent OCPs. Hence, they warm-start an SQP algorithm closely to the optimal solution and only perform a single SQP-iteration per time step. The RTI scheme is expected to converge to a locally optimal solution over multiple time steps, while the intermediate control inputs are already being applied each time step. They argue that, under some assumptions, the solution provided by the RTI scheme will follow the global solution of
the NMPC problem. The required assumptions are the following: (1)~the RTI scheme is warm-started at the global optimum, (2)~the sampling frequency is sufficiently high, (3)~there are no jumps in the reference and the state, (4)~the OCP underlying the NMPC problem fulfills Second-Order Sufficient Conditions for
every feasible initial condition, and (5)~the global optimum depends continuously on the initial state and reference~\cite{diehl_rti}.

\subsection{Accounting for Computational Delay}
% Related work on specifically ASAP-MPC, not on MPC algorithms
Instead of trying to reach a fixed (high) control rate, an alternative strategy is to allow more time to the NLP solver. Zavala et al.~\cite{Zavala2008} % moet dit niet "V. M. Zavala and L. T. Biegler. The advanced step NMPC controller: Optimality, stability and robustness. Automatica, 45:86–93, 2009.", of "The Advanced Step Real Time Iteration for NMPC" van Nurkanovic,.., Diehl zijn? in Zavala2008 staat nergens advanced step, en
detail an advanced-step NMPC algorithm based on the previously discussed RTI scheme. A fully converged solution to the NMPC problem is calculated during multiple control intervals, starting from an estimated future system state. Intermediate control inputs arise as an approximation of the fully converged solution but with perturbed initial conditions equal to the measured state. Using NLP sensitivity concepts, these intermediate control inputs are computed two orders of magnitude faster than the fully converged solution.

A second option is to explicitly account for the computational delay in the MPC scheme, as suggested by Findeisen et al.~\cite{FINDEISEN2004427}. However, lowering the planning and control rate for control systems with uncertainty makes it impossible to stabilise unstable systems, to reject high-frequency disturbances or react to environmental changes, and to cope with plant-model mismatch. In superposition to the feedforward inputs of an NMPC scheme, a faster linear feedback controller can be added, as done by Neunert et al.~\cite{Neunert2016} on a drone and a ballbot, and later by Grandia et al.~\cite{grandia2019} under the name of Feedback MPC on a quadruped robot. The Sequential Linear Quadratic algorithm used by both works, based on Dynamic Differential Programming, differs from conventional MPC in that the solution of the OCP consists of an optimised control sequence and state-dependent feedback policy. This policy is derived using a linearisation of the nonlinear problem around the optimal solution. The state feedback policy deals with high-frequency disturbances, while low-frequency disturbances are handled by the MPC updates. This hierarchical structure eliminates the sole dependence on the MPC to compensate all disturbances.
\section{Methodology}
% \section{methodology}
\label{section:ASAP}

To properly explain the working principle of ASAP-MPC, Sections~\ref{sec:furMPC} and~\ref{sec:lurMPC} first review how the classical MPC scheme, as discussed in~\cite{rawlings}, treats online updates, and how the approach proposed by Findeisen et~al. in~\cite{FINDEISEN2004427} copes with computational delay. These sections also list the assumptions made by both techniques. We will refer to the classical MPC scheme as Full Update Rate (FUR-) MPC and to the approach by Findeisen et~al. as Low Update Rate (LUR-) MPC. Next, Section~\ref{sec:traj_jump} details the trajectory jumping problem which can occur by carelessly applying the solution proposed by Neunert et~al.~\cite{Neunert2016}. Then, Section~\ref{sec:ASAP_MPC} presents our solution and finally, Section~\ref{sec:remarks} gives some concluding remarks. 
%Fig.~\ref{fig:classical_mpc} to Fig.~\ref{fig:asap_mpc} show the following schemes for the case of plant-model mismatch. 

\begin{figure*}
\begin{center}
\includegraphics[width=\linewidth]{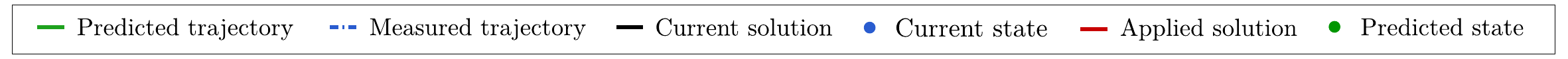}
\par\bigskip
\includegraphics[width=\linewidth]{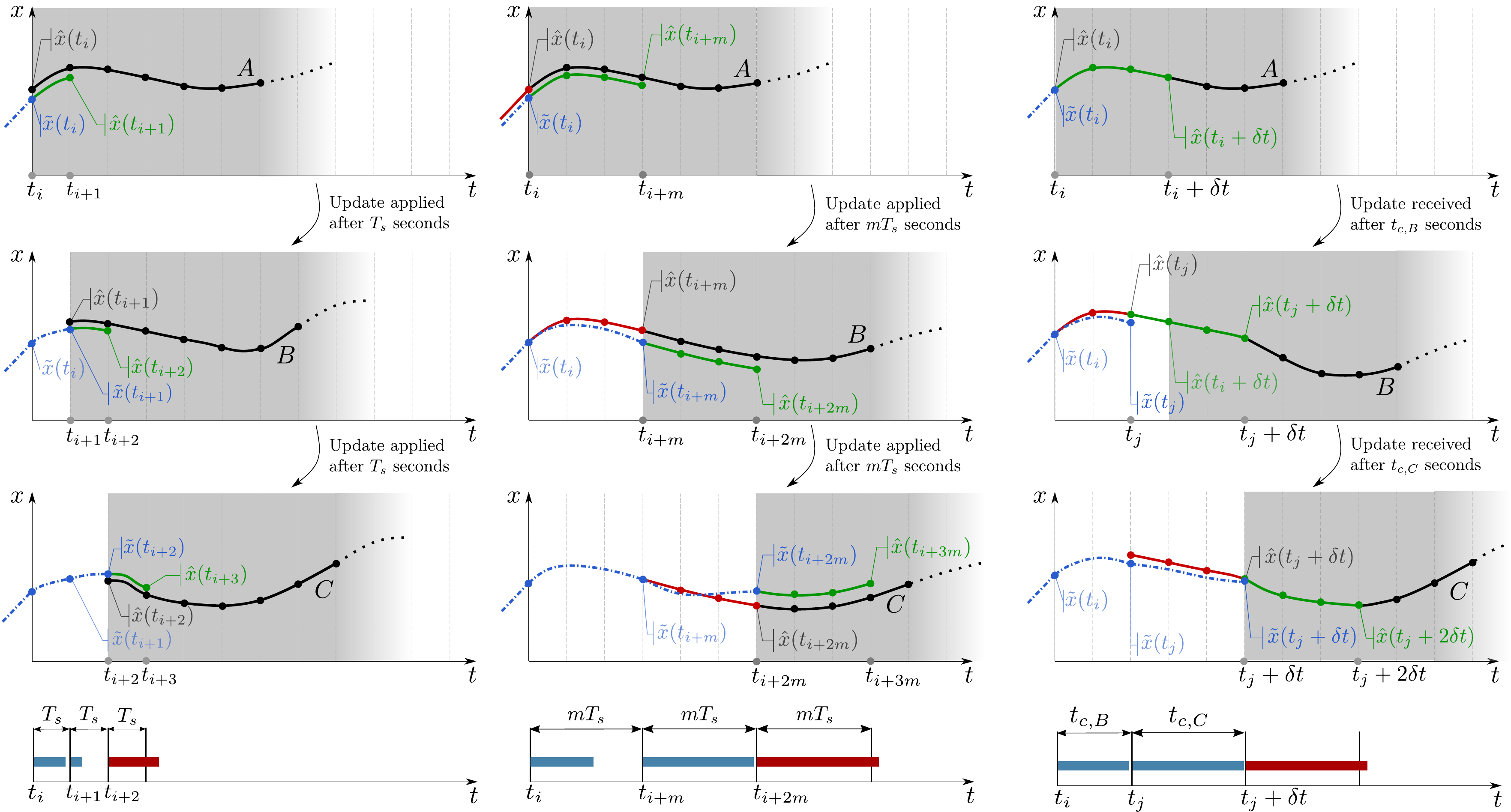}
\caption{Working principles of FUR-MPC (left), LUR-MPC (middle), and ASAP-MPC (right) and their corresponding timing diagrams (bottom row).}
\label{fig:mpc_working_principles}
\end{center}
\end{figure*}

Figure~\ref{fig:mpc_working_principles} illustrates the working principles for FUR-MPC, LUR-MPC and ASAP-MPC and highlights their differences. Each column shows the evolution of a 1D state together with the measured, planned and executed trajectories known at discrete time instances. A measurement of a state is denoted with a tilde ($\Tilde{x}$) and a prediction of a future state is denoted with a hat ($\hat{x}$).
% $t_{i}, t_{i+1}, t_{i+2}$.

% \begin{figure*}
% \begin{center}
% \includegraphics[width=\linewidth]{figures/legend_mpc_figures.pdf}
% \par\bigskip
% \begin{minipage}[b]{0.33333\textwidth}
% \includegraphics[width=\linewidth]{figures/classical_mpc.pdf}
% \caption{Classical MPC}
% \label{fig:classical_mpc}
% \end{minipage}%
% \begin{minipage}[b]{0.33333\textwidth}
% \includegraphics[width=\linewidth]{figures/findeisen_mpc.pdf}
% \caption{Findeisen MPC}
% \label{fig:findeisen_mpc}
% \end{minipage}%
% \begin{minipage}[b]{0.33333\textwidth}
% \includegraphics[width=\linewidth]{figures/asap_mpc.pdf}
% \caption{ASAP-MPC}
% \label{fig:asap_mpc}
% \end{minipage}
% \end{center}
% \end{figure*}

\subsection{Full Update Rate MPC}\label{sec:furMPC}
FUR-MPC schemes periodically compute optimal trajectories $x(t)$ with corresponding control inputs $u(t)$ over a future time horizon $T_{h}$. Only the first control input of each trajectory is applied for $T_s$ seconds, where $T_s$ is the sampling time. At the end of this interval, the current state $\tilde{x}(t_i + T_s)$ is measured and used as the starting point for the next trajectory and the whole procedure is repeated. A common theoretical assumption is that the new trajectory is immediately available after the state measurement. This assumption is always violated in practice. Even finding a solution within $T_s$ is challenging. Practical implementations are considered robust against this violation if $T_s$ is small enough. In the previously mentioned RTI scheme, most of the computations are executed as preparation before the state measurement and only a small, cheap correction is performed after the state measurement, yielding a very limited computational delay. 

% Figure~\ref{fig:classical_mpc} depicts the working principle of a Full Update Rate MPC scheme. At $t_{i}$, the current state {\color{darkblue} $\bar{x}(t_{i})$} is measured and leads to solution A. Applying the first input over a single control interval spanning $T_s$ seconds,  leads to an expected trajectory (\sampleline{color=green, thick, dashdotted}) up to {\color{red} $\hat{x}(t_{i + m})$}, the estimated next state. Due to plant-model mismatch, the actual system's state will be {\color{darkblue} $\bar{x}(t_{j})$} instead of {\color{red} $\hat{x}(t_{i + m})$}. Using this new measurement causes a corrective solution B (\sampleline{black, thick, solid}) compared to A (\sampleline{color=darkgray, thick, dashed}). The Full Update Rate MPC schemes repeat this procedure each control interval until a termination criterion for the task at hand is reached. The fourth part of Figure~\ref{fig:classical_mpc} depicts the MPC scheme's timing diagram. Within a single control interval, a new trajectory must be calculated. If this requirement is not met, a non-converged solution of the OCP is applied to the system which might violate constraints such as obstacle avoidance, actuator limits or might not correspond to the modelled nonlinear system dynamics. \\

The left column of Figure~\ref{fig:mpc_working_principles} outlines the working principle of a FUR-MPC scheme, using a future state estimator to already cope with a computational delay smaller than $T_{s}$. Assume that at $t_{i}$, a solution~$A$ from a previous computation is available. The current state $\Tilde{x}(t_{i})$ is measured and is used to predict the state at $t_{i+1}$ using the first control input from solution~$A$. This predicted state $\hat{x}(t_{i+1})$ is used as the initial condition for solution~$B$. This solution must be available one time step later. The first control input $u(t_{i})$ from solution~$A$ is applied to the system in the meantime spanning $T_s$ seconds. At $t_{i+1}$, in the second figure, the actual (measured) state has evolved to $\Tilde{x}(t_{i+1})$ instead of $\hat{x}(t_{i+1})$ due to plant-model mismatch, and solution~$B$ becomes available. This procedure is now repeated, with an estimated $\hat{x}(t_{i+2})$, to start the computation for solution~$C$, visualised in the third graph. The bottom part of Figure~\ref{fig:mpc_working_principles} depicts a possible scenario for the MPC scheme's timing diagram. The blue bar indicates the actual computation time, and the width of the intervals is the controller's sampling time $T_s$. Note that if the computation takes longer than the sampling time, the case with the red bar, the MPC update fails. % Within each single control interval, a new trajectory must be computed.
The working principle is also shown in Algorithm~\ref{alg:FUR-MPC}.

% Different lines:
% \begin{itemize}
%     \item Expected: \sampleline{color=green, thick, dashdotted}
%     \item Measured: \sampleline{color=lightblue, thick, densely dashed} 
%     \item Past: \sampleline{color=lightgray, thick, densely dashdotted}
%     \item Planned: \sampleline{color=darkgray, thick, dashed}
%     \item Current: \splitline{red}{black}{thick}{solid}%, thick, solid} %\sampleline{red, thick, solid} \sampleline{black, thick, solid} 
% \end{itemize}

 % Only the first control input is applied, as it is assumed that within the sampling time $T_s$ a new trajectory is calculated for the horizon receded by one sample. In fact, it is assumed that there is no computation delay, and the OCP solution is immediately available. However, as mentioned before, computing a solution to a complex OCP within the sampling time $T_s$ is very challenging. Moreover, the assumption of zero computation delay is never satisfied. In practice, the MPC scheme is assumed to be robust against the computation delay if the sampling time is small enough.

\subsection{Low Update Rate MPC}\label{sec:lurMPC}
The LUR-MPC scheme is a strategy that incorporates the finite computational delay that arises from computing a new trajectory and is larger than $T_s$. The delay can span multiple control intervals, in contrast with the FUR-MPC scheme where a new trajectory has to be found within a single control interval. It assumes that the maximum computation time, expressed as a positive, whole number of the sampling time $mT_s$, is known~\cite{FINDEISEN2004427}.

%\red{check legende dot streep + onderscheid planned/current zwart en donker grijs}
%Figure~\ref{fig:findeisen_mpc} visualises the principle of the Low Update Rate MPC algorithm. At $t_{i}$, the current state {\color{darkblue} $\Tilde{x}(t_{i})$} is measured. Firstly, solution A starts from this initial state. Since the maximum delay is known, the next solution (B) will start from the expected system state {\color{red} $\hat{x}(t_{i + m})$}. Henceforth, the inputs from solution A are applied open-loop up to $t_{i + m}$ with $m=\delta_{c}$, leading to an expected trajectory for the system (\sampleline{color=green, thick, dashdotted}). Due to plant-model mismatch, the trajectory that the system will actually follow (\sampleline{color=lightblue, thick, densely dashed}) deviates from the expected trajectory \red{hier is iets mis}(\sampleline{color=lightblue, thick, densely dashed}). Consequently, solution B starts from a wrong state. {\color{red} How do we phrase this correctly = sort of drifting problem} Same as for the Full Update Rate MPC scheme, this procedure is repeated until a termination criterion is reached. The advantage of \red{sampled-data NMPC??} is that more complex problems can be solved due to the foreseen margin of computational delay. In real mechatronic applications, plant-model mismatch will always exist. Hence, one drawback of this sampled-data NMPC scheme is the fact that the calculated inputs are applied in open-loop, leading to the drifting problem.

The middle column of Figure~\ref{fig:mpc_working_principles} visualises the working principle of the LUR-MPC algorithm. At $t_{i}$, a solution~$A$ from a previous computation is available. The current state $\Tilde{x}(t_{i})$ is measured and is used as an initial condition for solution~$B$. This solution must be available $m$ time steps later. It starts with the first $m$ control inputs from solution~$A$, as they are fixed in the optimisation problem. These $m$ control inputs from solution~$A$ are applied to the system in open loop over a period of $mT_s$ seconds. At $t_{i+m}$, in the second figure, the actual (measured) state has evolved to $\Tilde{x}(t_{i+m})$ instead of $\hat{x}(t_{i+m})$ due to plant-model mismatch, and solution~$B$ becomes available. This procedure is now repeated to start the computation for solution~$C$, visualised in the third figure. The lowest figure depicts this MPC scheme's timing diagram. Within a fixed number of control intervals, a new trajectory must be computed. Note that due to plant-model mismatch, the measured trajectory deviates from the expected trajectory. Hence, one drawback of this NMPC scheme is drift as the computed control inputs are applied in open-loop over $mT_s$ seconds.
Algorithm~\ref{alg:LUR-MPC} further details the working principle.
% On the right in Fig.~\ref{fig:classic_and_Findeisen_updates}, this is shown for a computation delay of $\Delta k$ samples. From the computed trajectory, not only one, but $\Delta k$ control input samples are applied (highlighted in red). To cope with the computation delay inside the OCP, it is already taken into account in the OCP formulation that the first $\Delta k$ samples cannot be taken from the new solution, as it will only become available after the computation delay has passed. Instead, the control inputs that are applied during this computation time are sampled from the previous solution. In the OCP, the first $\Delta k$ control inputs are therefore constrained to be equal to these samples from the previous solution (highlighted in blue). This is equivalent to simulating the estimated future state $\hat{\vec{x}}(t_{k+2\Delta k})$ of the robot using the planned control inputs from the currently available solution, starting from the current measured state $\vec{x}(t_{k+\Delta k})$.

% \begin{figure}
%     \centering
%     \includegraphics[width=\linewidth]{figures/concept_classic_and_Findeisen_mpc.pdf}
%     \caption{Working principle of the update strategies, illustrated both in time and in space, for both classic MPC (left) and Findeisen (right).\red{TO BE UPDATED WITH DRIES' NEW FIGURES}}
%     \label{fig:classic_and_Findeisen_updates}
% \end{figure}

\subsection{The Trajectory Jumping Phenomenon}
\label{sec:traj_jump}
\begin{figure}
    \centering
    \includegraphics[width=0.9\linewidth]{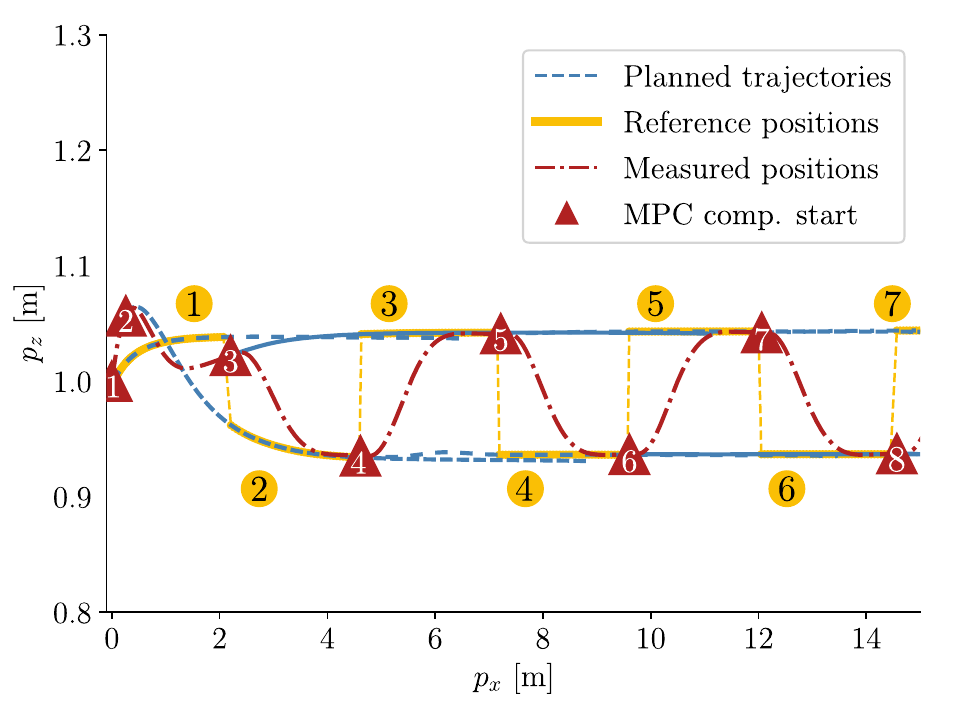}
    \caption{Illustration of the trajectory jumping phenomenon for an aerial robot flying forwards in free space, occurring when naively implementing the methodology from Neunert et al.~\cite{Neunert2016}.}
    \label{fig:traj_jumping}
\end{figure}

As mentioned in Section~\ref{section:introduction}, Neunert et~al.~\cite{Neunert2016} propose to add a tight, faster linear feedback controller to the MPC scheme to eliminate drift, such that this combined control system is capable of realising accurate and agile motion. The MPC is responsible for re-planning and dealing with low-frequency disturbances, while the feedback controller addresses the high-frequency disturbances, stabilisation, and plant-model mismatch. However, a trajectory jumping phenomenon can occur when Neunert's strategy is naively implemented. This phenomenon is illustrated in Figure~\ref{fig:traj_jumping} for an aerial robot flying forwards in a free space. From standstill, the first trajectory~(\sampleline{steelblue, thick, dashed}) is computed and tracked by the state feedback controller. The reference for this controller is shown in yellow~(\sampleline{jump_yellow, ultra thick}). After the update period of $mT_s$ seconds, the robot's state is measured~(\redtriangle) and using this measurement as initial point, a new solution to the OCP is computed. $m$ is chosen slightly exaggerated to highlight the effect. Due to possible drift, this measured state might not match with its expected state on the reference trajectory. As the new solution does not arrive instantaneously, the feedback controller will force the robot back to the currently available solution~(\sampleline{jump_yellow, ultra thick}). When the new solution~(\sampleline{steelblue, thick, densely dashed}) becomes available, the reference trajectory of the feedback controller~(\sampleline{jump_yellow, ultra thick}) suddenly shifts to this new solution, forcing the feedback controller to severely intervene. The measurement that determines the initial state of the next trajectory is taken right before this intervention (shortly after the time at which the solution became available), which causes the next solution to be very different from the current one. The same procedure keeps on repeating, ultimately leading to the robot travelling along a zig-zag trajectory~(\sampleline{firebrick, thick, dashdotted}) between two sets of trajectories. Clearly, this is sub-optimal and should be avoided. Note that the phenomenon is presented here in an exaggerated way for visual clarity, but also at lower amplitudes this behaviour is undesired.

% At some point, the robot state is measured (\sampleline{red, thick, dashdotted}), and a trajectory is computed with that state as initial condition. If the robot, for whatever reason, deviates from the trajectory, then the next measured state (dark grey) will not lie on the planned trajectory (exaggerated in the figure for visual clarity). During the computation of the new OCP solution (black trajectory), the feedback controller urges the robot back to the reference, until the new solution becomes available, and the reference shifts to the black trajectory. Now the same sequence occurs, and in the end the resulting traveled trajectory is a zig-zag between two sets of trajectories. Obviously this is sub-optimal, and to be avoided.

\begin{figure}
    \centering
    \includegraphics[width=0.9\linewidth]{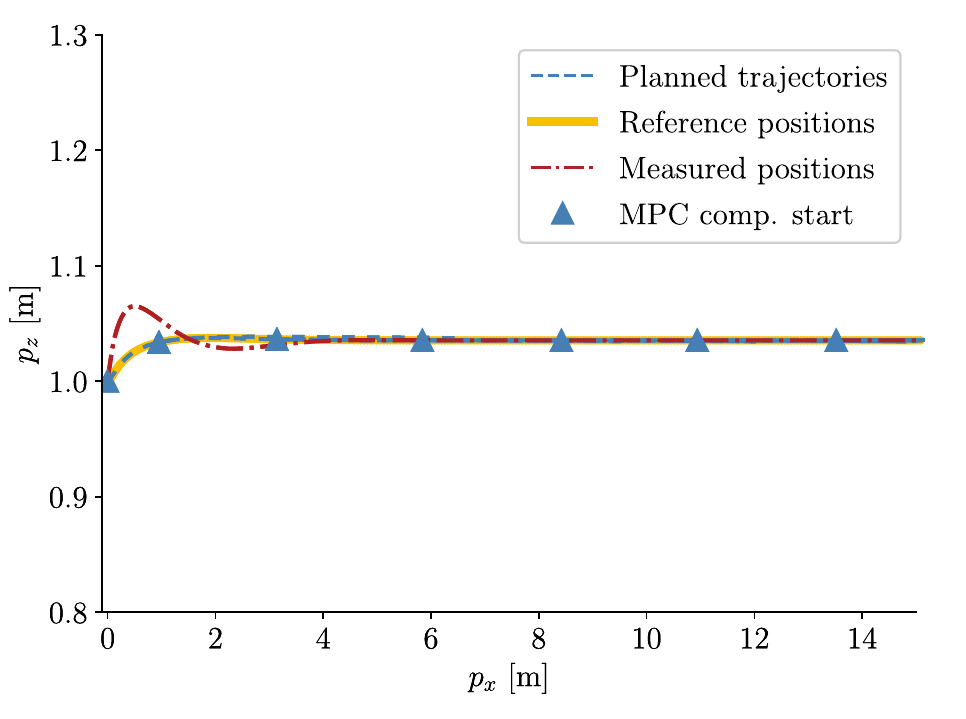}
    \caption{ASAP-MPC: a solution to the trajectory jumping problem illustrated in Fig.~\ref{fig:traj_jumping}.}
    \label{fig:traj_jumping_2}
\end{figure}

\begin{table}
\centering
\caption{Summary of some properties for the discussed MPC schemes.}
\label{tab:prop_summ}
    \begin{tabular}{cccc}
        \toprule
        \textbf{MPC Scheme} & FUR-MPC & LUR-MPC & ASAP-MPC \\
        \midrule
        Update interval & $T_{s}$ & $mT_{s}$ & variable \\
        Extra feedback controller & $\times$ & $\times$ & \checkmark \\
        Fast disturbance rejection & \checkmark & $\times$ & \checkmark \\
        \bottomrule
    \end{tabular}
\end{table}

\subsection{ASAP-MPC Scheme}\label{sec:ASAP_MPC}
As a solution to the aforementioned problems originating from the approaches by Neunert and Findeisen, the authors propose ASAP-MPC. This strategy follows the reasoning of Neunert to add linear feedback control to the NMPC. On the other hand, it resembles the approach of Findeisen as it estimates the future state of the robot, and optimises the trajectory starting there. However, it introduces two differences with Findeisen's approach. Firstly, instead of adding (and fixing) the first $m$ control inputs to the OCP from the previous solution, they are discarded from the OCP and the initial condition of the OCP is the predicted future state itself. This initial condition is sampled from the trajectory of the previous solution, as this is the best guess of where the robot will be, because the addition of the linear state feedback controller urges the robot back to this reference. A newly computed trajectory is stitched to the previous trajectory at this future state. The remainder of the previous trajectory is discarded. Secondly, there is no reason to wait until the full $m$ samples have passed to start a new computation. ASAP-MPC starts the next computation as soon as the previous one finishes. 

The third column of Figure~\ref{fig:mpc_working_principles} visualises the working principle of ASAP-MPC. In order to emphasise the asynchronous characteristics of this method, we no longer use a discrete time notation ($t_{i+1}$), but a continuous representation ($t_{i}+t_\text{comp}$). Again assume that at $t_{i}$ a solution~$A$ from a previous computation step is available. The expected state at $t_{i}+\delta t$, with $\delta t = mT_s$, is taken from solution~$A$ directly and not based on a forward simulation starting at the current state measurement $\Tilde{x}(t_{i})$. Note that this relies heavily on the assumption that the state feedback controller is capable of keeping the state close to the trajectory of solution~$A$. The predicted state $\hat{x}(t_{i}+\delta t)$ is now used as initial state for solution~$B$. During the computation of the new trajectory, the computed state trajectory is tracked by the state feedback controller. As soon as solution~$B$ becomes available after the computation time $t_\text{c,B}$, with $t_\text{c,B} < \delta t$, it is directly stitched to the previous solution at $t_{i}+\delta t$, which will still be a future time instance. The procedure repeats immediately at least as fast as, but most often faster compared to LUR-MPC's update rate. At $t_{i} + t_\text{c,B}$, in the second figure, the actual (measured) state has evolved to $\Tilde{x}(t_{i} + t_\text{c,B})$. Again, this value is not used to start the procedure for solution~$C$, but instead it starts from the estimated state $\hat{x}(t_{i}+t_\text{c,B}+\delta t)$ resulting from solution~$B$. The third figure shows the same procedure with a remarkably longer computation time than the first two situations, but still smaller than $\delta t$. Remark that there are no jumps in the reference trajectory for the state feedback controller as a result of this stitching approach. The bottom part of the figure depicts this MPC scheme's timing diagram. Within a fixed number of control intervals, a new trajectory must be computed, but if the solution becomes available sooner, the procedure repeats as soon as possible. Note that the predicted state always lies $m$ samples in the future ($\delta t$), selected relative to the time at which the previous computation finishes, e.g. the very first one being $t_{i}+t_\text{comp}+\delta t$. Starting a new computation earlier realises a faster reaction to environmental changes in the best case scenario, while still allowing for more time to compute the solution in the worst case scenario.
The working principle is also shown in Algorithm~\ref{alg:ASAP-MPC}. Note that it also includes an application-specific safety precaution in case no solution is found within the available time. An implementation of such an emergency reaction is proposed in \cite{drone2023}.

Figure~\ref{fig:traj_jumping_2} shows how this approach solves the trajectory jumping phenomenon. As the OCP starts from predicted states on the previous trajectory~(\bluetriangle), instead of starting from the measured positions~(\redtriangle), the resulting reference trajectory is smooth.

% Firstly, instead of adding the first $\Delta k$ control inputs to the OCP, the OCP is directly computed with the estimated future state as the initial condition, hence reducing the size and complexity of the OCP.\\
% Secondly, due to the addition of the linear feedback controller that urges the robot back to the reference, the estimated future state is not the one simulated using the control input commands from the previous solution, but instead a point sampled from the trajectory of the previous solution. A new trajectory is always stitched to the old trajectory at this sampled point. The remainder of the old trajectory is discarded.\\
% Finally, there is no reason to wait until the full $\Delta k$ samples have passed to start a new computation. ASAP MPC starts the next computation as soon as the previous one finishes. The update state that lies $\Delta k$ samples in the future is always selected relative to the time at which the previous computation finishes. Starting a new computation as soon as possible has the advantage of incorporating environmental changes faster in the best case scenario, while still allowing for more time to compute the solution in the worst case scenario.

% \begin{figure}
%     \centering
%     \includegraphics[width=\linewidth]{figures/asap_mpc.png}
%     \caption{Working principle of the update strategy in ASAP MPC, illustrated both in space and time.}
%     \label{fig:ASAP_updates}
% \end{figure}
\vspace{-2ex}
\subsection{Practical Remarks}\label{sec:remarks}
The first important remark is that the strategy to sample the initial condition of the OCP from the current reference trajectory rather than measuring it implies that the NMPC incorporates no feedback on the robot state itself. This is handled by a separate controller combining state feedback and feedforward control inputs from the NMPC solutions to realise accurate trajectory tracking. The NMPC does incorporate feedback on the perceived environment, e.g. obstacles and boundaries of free space.% \red{the experiments show the bounds on the tracking performance}

Secondly, the choice of $m$ is an important design trade-off. On the one hand, $m$ should be taken small enough, such that the motion planning algorithm can adjust the path in the near future. A high $m$ fixes the planned state  too far into the future, such that newly appearing obstacles or other occurring environmental changes might not be corrected for in time by the trajectory planner. On the other hand, $m$ should be taken large enough, such that in the worst case scenario with respect to computation time, it grants the solver sufficient time to find a solution before the robot reaches the update point.

Thirdly, even though the presented algorithm allows for more time to find a solution, rapid computations of solutions are still desirable. Swift updates take environmental effects more rapidly into account, leading to a higher chance of success in the task at hand. Therefore, taking care in designing the OCP and finding a suitable solver for the application is still beneficial for the performance of the ASAP-MPC scheme.

\begin{algorithm}%\small
    \caption{FUR-MPC scheme}
    \label{alg:FUR-MPC}
    \begin{algorithmic}[1]
        \Require $T_s$
        \State $i,  u_0(t_0) \gets 0$
        \Repeat $\:$every $T_s$
        \State apply $u_{i}(t_{i})$
        \State measure $\tilde{x}(t_i)$
        \State predict $\hat{x}(t_{i+1})$ from $\tilde{x}(t_i)$
        \State $u_{i+1}, x_{i+1} \gets \texttt{solve\_FUR\_ocp}(\hat{x}(t_{i+1}))$
        % \State $u_{i+1}, x_{i+1} \gets \text{get\textunderscore ocp\textunderscore sol}(\hat{x}(t_{i+1}))$
        \State $i \gets i + 1$
        \Until{task finished}
    \end{algorithmic}
\end{algorithm}
% \vspace{-4ex}

\begin{algorithm}%\small
    \caption{LUR-MPC scheme}
    \label{alg:LUR-MPC}
    \begin{algorithmic}[1]
        \Require $T_s$, $m$, $\delta t = m T_s$
        \State $i, k, \ell \gets 0$
        \State measure $\tilde{x}(t_0)$
        % \State predict $\hat{x}(t_0)$ from $\tilde{x}(t_0)$
        \State $u_0, x_0 \gets \texttt{solve\_LUR\_ocp}(\tilde{x}(t_0)))$

        % \State measure $\tilde{x}(t_0)$
        % \State $\text{start\textunderscore solve\textunderscore ocp}(\tilde{x}(t_0))$
        % \State $u_0, x_0 \gets \text{get\textunderscore ocp\textunderscore sol}(\hat{x}(t_0))$
        \Repeat $\:$every $\delta t$
        \State measure $\tilde{x}(t_i)$
        \State \begin{varwidth}[t]{\linewidth}
            start $\texttt{solve\_LUR\_ocp}(\tilde{x}(t_i), u_{\ell-1}(t_i),$ \par
            \hskip\algorithmicindent $u_{\ell-1}(t_{i+1}), ..., u_{\ell-1}(t_{i+m-1}))$
        \end{varwidth}\vspace{1mm}
        \While{$k \in \left[ 0, m-1\right]$} every $T_s$
        \State apply $u_\ell(t_{i+k})$
        \State $k \gets k+1$
        \EndWhile
        \State \begin{varwidth}[t]{\linewidth}
            $u_\ell, x_\ell \gets \texttt{get\_ocp\_sol()}$
        \end{varwidth}\vspace{1mm}
        \State $i \gets i + m$
        \State $\ell \gets \ell + 1$
        \State $k \gets 0$
        \Until{task finished}
    \end{algorithmic}
\end{algorithm}
% \vspace{-4ex}

\begin{algorithm}%\small
    \caption{ASAP-MPC scheme}
    \label{alg:ASAP-MPC}
    \begin{algorithmic}[1]
        \Require $T_s$, $m$, $K_{fb}$, $\delta t = m T_s$
        \State $i, k, \ell \gets 0$
        \State measure $\tilde{x}(t_0)$
        % \State predict $\hat{x}(t_0)$ from $\tilde{x}(t_0)$
        \State $u_0, x_0 \gets \texttt{solve\_ASAP\_ocp}(\tilde{x}(t_0))$
        % \State $u_0, x_0 \gets \text{get\textunderscore ocp\textunderscore sol}(\hat{x}(t_0))$
        \Repeat $\:$
        \State predict $x_\ell(t_i + \delta t)$
        \State $\text{start } \texttt{solve\_ASAP\_ocp}(\hat{x}(t_i + \delta t))$
        \While{$k \in \left[ 0, m-1\right]$ and no solution} every $T_s$
        \State $t_k \gets t_{i} + k T_s$
        \State measure $\tilde{x}(t_k)$
        \State $u_k = u_\ell(t_k) + K_{fb}(x_{\ell}(t_k) - \tilde{x}(t_k))$
        \State apply $u_k$
        \State $k \gets k+1$
        \EndWhile
        \If{no solution}
        \State \text{do }$\texttt{emergency\_reaction}()$
        \State \textbf{break}
        \Else
        \State $u_\ell, x_\ell \gets \texttt{get\_ocp\_sol}()$
        \State $t_{\ell+1} \gets t_\ell + t_{c,\ell}$
        \State $\ell \gets \ell+1$
        \EndIf
        \State $i \gets i + m$
        \State $k \gets 0$
        \Until{task finished}
    \end{algorithmic}
\end{algorithm}
% !TeX root = main.tex
%\section{Experiment setups}
\section{Experimental Validation}
\label{section:experiment_setups}

This section covers the two applications that are presented as use cases of the ASAP-MPC methodology, extensions with respect to the work in two previous papers on the same applications, the conducted experiments to validate the methodology, and the validation metrics. Finally it presents a comparison between the usage of ASAP-MPC and the RTI scheme.

\subsection{Use Cases}

\begin{table*}
\caption{ODEs describing the kinematics and dynamics of the quadrotor drone and truck-trailer AMR.}
\centering
\begin{tabular}{cc} 
\toprule
Drone & Truck-trailer AMR\\
\midrule
    \begin{minipage}[t]{0.45\linewidth}
        \begin{equation*}
        \begin{aligned}
    \vec{x} &= \begin{bmatrix} p_x & p_y & p_z & v_x & v_y & v_z & \phi & \theta & \psi \end{bmatrix}\transp\\
    \vec{u} &= \begin{bmatrix}p & q & r & a_t \end{bmatrix}\transp\\
                        % \begin{bmatrix} \vec{v} \\
                        % \matrix{R}\left[0, 0, a_t \right]\transp -\vec{g} \\
                        % \matrix{S}^{-1} \vec{\omega} \end{bmatrix}  \\
    \dot{\vec{x}} &= \vec{f}(\vec{x}, \vec{u})\\
                  &= \begin{bmatrix} \vec{v} \\
                        \matrix{R}\begin{bmatrix}0 & 0 & a_t \end{bmatrix}\transp -\vec{g} \\
                        \matrix{S}^{-1} \vec{\omega} \end{bmatrix} 
                        % \\
        \end{aligned}
        \label{eq:dynamics_model}
        \end{equation*}
where
    \begin{equation*}
        \vec{\omega} = \begin{bmatrix} p & q & r\end{bmatrix}\transp = \matrix{S}\begin{bmatrix}\dot{\phi} & \dot{\theta} & \dot{\psi}\end{bmatrix}\transp
    \end{equation*}
    \vspace{0.35cm}
    \begin{equation*}
    \begin{aligned}
    p_x, p_y, p_z\quad & \textrm{position}\\
    v_x, v_y, v_z\quad & \textrm{velocity}\\
    \phi, \theta, \psi\quad & \textrm{attitude (roll-pitch-yaw Euler angles)}\\
    p, q, r\quad & \textrm{angular velocity}\\
    a_t\quad & \textrm{thrust acceleration / mass-normalized collective thrust}
    \end{aligned}
    \end{equation*}
    \end{minipage} &
    \begin{minipage}[t]{0.45\linewidth}
        \begin{equation*}
        \begin{aligned}
\vec{x} &= \begin{bmatrix} p_{x,1} & p_{y,1} & \theta_1 & \theta_0 \end{bmatrix}\transp\\
\vec{u} &= \begin{bmatrix} v_0 & \omega_0 \end{bmatrix}\transp\\
\dot{\vec{x}} &= \vec{f}(\vec{x}, \vec{u})\\
              &= \begin{bmatrix}
                  v_1 \cos\theta_1\\
                  v_1 \sin\theta_1\\
                  \frac{v_0}{L_1}\sin(\theta_{0} - \theta_{1}) - \frac{M_0}{L_1}\omega_0\cos(\theta_{0} - \theta_{1})\\
                  \omega_0
              \end{bmatrix}
        \end{aligned}
        \end{equation*}
where $v_1(t)$ is given by
        \begin{equation*}
v_1 = v_0 \cos(\theta_{0} - \theta_{1}) + M_0\omega_0\sin(\theta_{0} - \theta_{1})
        \end{equation*}
\begin{equation*}
\begin{aligned} 
p_{x,1}, p_{y,1}\quad & \textrm{trailer position}\\
v_0, v_1\quad &\textrm{longitudinal velocity of the truck and the trailer}\\ 
\omega_0\quad &\textrm{angular velocity of the truck}\\
\theta_{0}, \theta_{1}\quad &\textrm{orientation of the truck and trailer, detailed in~\cite{tt2023}}\\
L_1\quad &\textrm{distance between the steering wheel/hitching point and}\\
\quad & \textrm{the axle center}\\
M_0\quad &\textrm{distance between the axle center and the hitching point}
\end{aligned}
\end{equation*}
    \end{minipage} \\
    \bottomrule
\end{tabular}
\label{tab:models}
\end{table*}

% p_{x,1}, p_{y,1}: &Trailer position
% v_0, v_1: &\textrm{Longitudinal velocity of the truck and the trailer}\\ 
% \omega_0: &\textrm{Angular velocity of the truck}\\
% \beta_{01}: &\textrm{Angle between the truck and the trailer}\\
% L_0, L_1: &\textrm{Distance between the steering wheel/hitching point and the axle center}\\
% M_0: \textrm{Distance between the axle center and the hitching point}

To validate the performance of the presented framework, it is implemented for two applications:
\begin{itemize}
    \item Drone navigation through a locally known, cluttered environment. The task is to travel a specified distance as quickly as possible, without colliding with both static and moving spherical obstacles of varying dimensions and velocities.
    \item Truck-trailer autonomous mobile robot (AMR) manoeuvring in confined spaces, where the vehicle performs a parking manoeuvre. The environment is represented as a sequence of freely accessible rectangles (or corridors) and constraints are defined to keep the vehicle(s) within these corridors. This approach avoids to define obstacle avoidance constraints for every single obstacle. A high-level planner provides a set of corridors to solve the current optimal control problem. 
\end{itemize}

Table~\ref{tab:models} shows the ordinary differential equations (ODEs) governing the kinematics and dynamics of the two applications, written as a function of the state vector $\vec{x}$ and the control input vector $\vec{u}$. Both applications are further detailed in two corresponding papers,~\cite{drone2023} and~\cite{tt2023}. 

\subsection{Extensions to Previous Works}
As mentioned in Section~\ref{section:introduction}, running complex algorithms onboard with limited available processing power within a certain time window is challenging. To showcase the presented framework's capability to run on-line and onboard, this paper extends the work in~\cite{drone2023} and~\cite{tt2023}, which feature ASAP-MPC with offboard computations, by conducting all experiments with \textit{onboard motion planning and control computations}. A commonly used embedded platform is mounted on the physical setup in the truck-trailer AMR case, and used in hardware-in-the-loop (HiL) simulations in the drone case. The embedded platform is an NVIDIA Jetson TX2 with Quad-Core ARM\textsuperscript{\textregistered} Cortex\textsuperscript{\textregistered}-A57 MPCore CPU with six cores with a frequency range of 345.6~MHz - 2.035~GHz and 7.67~GB~RAM and NVP model Max-N. \\ % It forms a good example of the algorithm's intended application since its CPU is less powerful than typical CPUs in portable computers. \\

Ipopt was found to be unable to compute solutions sufficiently fast on this onboard platform for the applications in \cite{tt2023} and \cite{drone2023}, due to the (dynamically) changing environment. As an improvement over earlier work, this work uses the Fatrop solver \cite{fatrop} instead of Ipopt for the OCPs. Fatrop is an optimal control problem solver that aims to be fast, solve a broad class of OCPs and achieve a high numerical robustness. By the integration of this solver, the trajectory update times could be significantly improved. This increased computational performance leads to better capabilities of responding to disturbances and dynamically changing environments. For the truck-trailer application, all inequality constraints are enforced with a penalty method to allow Fatrop to robustly find a solution at the corridor changes.%We found this was particularly important when deploying the proposed control scheme on onboard platforms. 
% The chosen rates for FUR and LUR-MPC are 50 Hz and 1.25 Hz, respectively. These correspond to the rate necessary to stabilise and control the drone (50 Hz) and the maximal time allowed to find a solution to the OCP (1.25 Hz).\\

\subsection{Experiment Setup}
\begin{figure}
    \begin{subfigure}[t]{0.24\textwidth}
        \centering
        \includegraphics[width=0.95\linewidth]{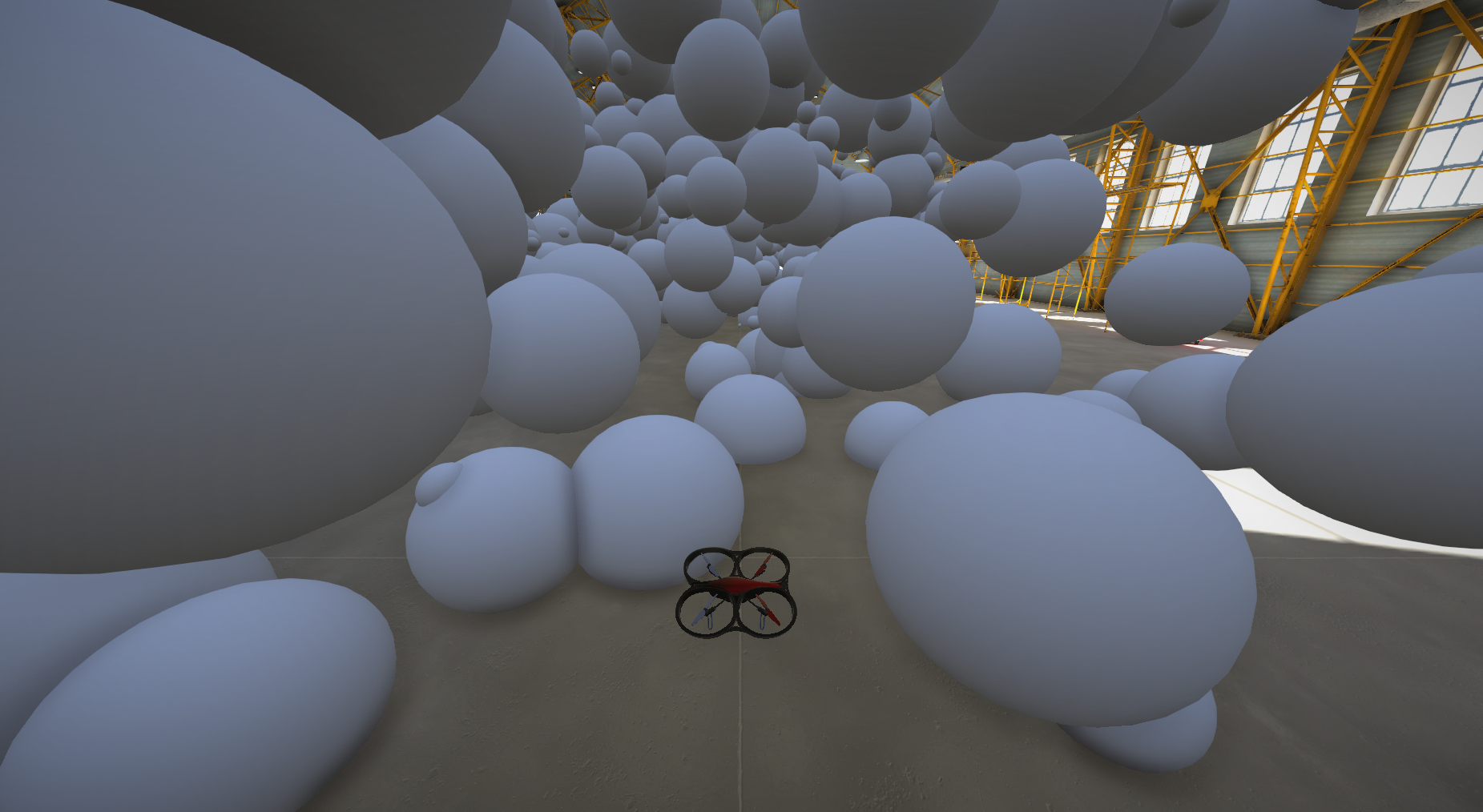}
        \caption{Flightmare environment}
        \label{fig:drone_nav_a}
    \end{subfigure}
    % \hspace*{\fill}
    \begin{subfigure}[t]{0.24\textwidth}
        \centering
        \includegraphics[width=0.95\linewidth]{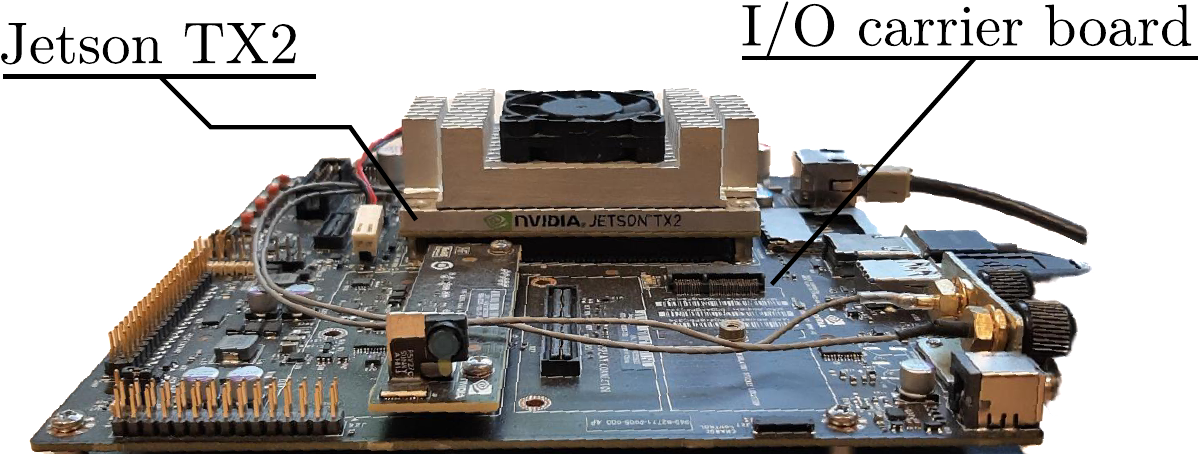}
        \caption{Jetson TX2 embedded platform.}
        \label{fig:drone_nav_b}
    \end{subfigure}
    \caption{Illustration of the used simulation environment and the onboard processor used in the drone application experiments.}
    \label{fig:drone_nav}
\end{figure}

\begin{figure}
    \begin{subfigure}[t]{0.24\textwidth}
        \centering
        \includegraphics[width=4.15cm,height=2.3cm]{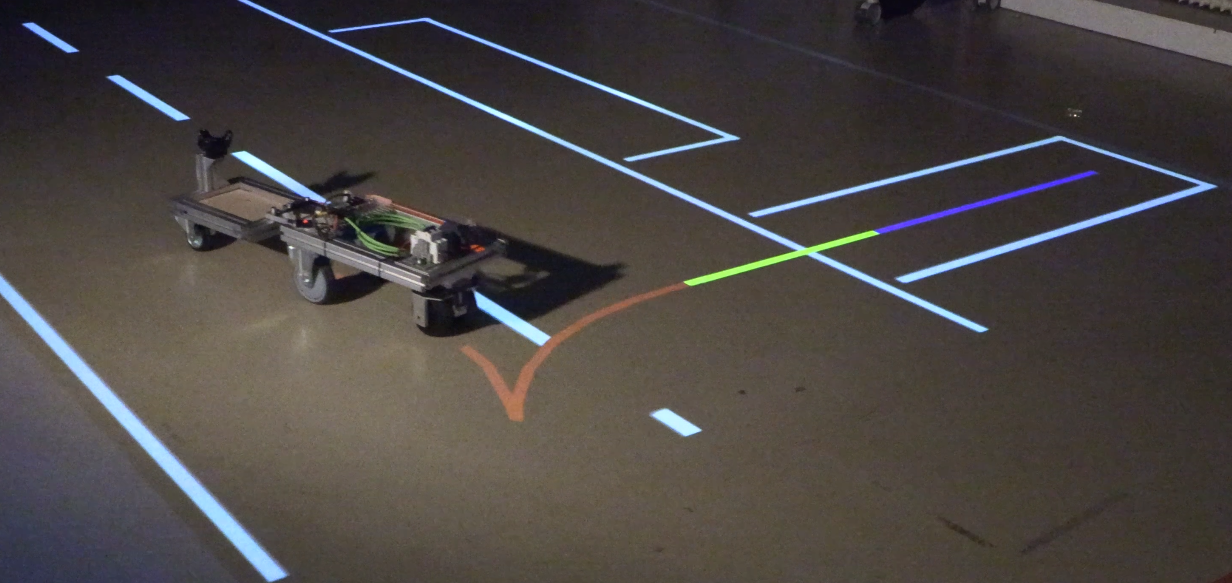} %width=\linewidth
        \caption{Truck-Trailer AMR manoeuvring}
        \label{fig:tt_man_a}
    \end{subfigure}
    % \hspace*{\fill}
    \begin{subfigure}[t]{0.24\textwidth}
        \centering
        \includegraphics[width=4.15cm,height=2.3cm]{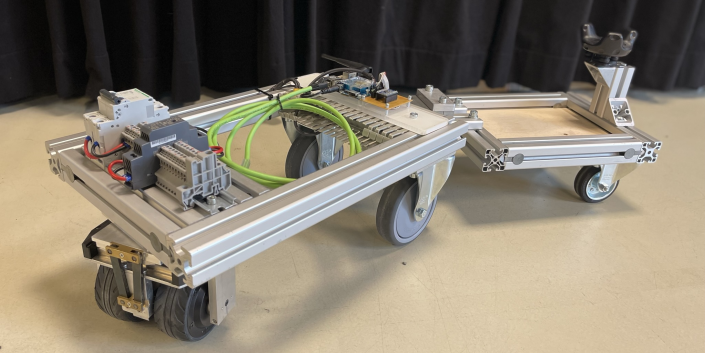} %width=0.47\linewidth
        \caption{Truck-Trailer AMR}
        \label{fig:tt_man_b}
    \end{subfigure}
    \caption{Illustration of the physical set-up used in the truck-trailer AMR experiments.}
    \label{fig:tt_man}
\end{figure}

The drone navigation experiments use the Flightmare simulator~\cite{song2020flightmare}, while the truck-trailer experiments are performed on a physical lab set-up, both shown in Figures~\ref{fig:drone_nav} and~\ref{fig:tt_man}. Flightmare accurately simulates drone dynamics while eliminating practical complications such as drone localisation, safety hazards, etc. Since the simulation is not part of the presented algorithm, it is running separately on a laptop.\\
% \red{with an Intel\textsuperscript{\textregistered} Core™ i7-10810U processor with twelve cores at 1.10GHz and 31.1 GB RAM. 
% MB: is het nodig om de specs van de laptop te vermelden? aangezien die eigenlijk geen invloed heeft op de prestaties van het gepresenteerde werk?} \\

Compared to the experiments conducted in~\cite{drone2023}, this paper only uses a limited set of environments from the DodgeDrone Challenge. Five easy-level environments and five medium-level environments are each run twice, resulting in a total of 3274 computed trajectories. 
For the truck-trailer AMR, this paper executes the perpendicular parking manoeuvre twelve times, resulting in a total of 6484 computed trajectories.
% Table: p-norm, x-norm, y-norm, z-norm

\subsection{Validation Metrics}
The following validation metrics are studied:
\begin{itemize}
    \item feedback error between reference trajectory and measured trajectory,
    \item computation times (wall time)
    % \item Comp{\color{red} Comparison with vanilla RTI algorithm}
\end{itemize}
These characteristics are chosen to validate the claims, discussed in Section~\ref{sec:ASAP_MPC}, that (1)~the on-trajectory future state chosen as the best estimated guess of the future robot's state is reached in close proximity and (2)~a framework able to deal with variable computational delay is necessary and desired for complex applications. The necessity of the presented framework is further discussed by comparing it to a widely used state-of-the-art online MPC methodology, the RTI scheme introduced in Section~\ref{section:related_work}. \\
% and (3) that {\color{red} a popular state-of-the-art online MPC methodology is not applicable to the presented applications / the ASAP-MPC framework outperforms a ...} . The RTI algorithm as introduced in Section~\ref{section:related_work} is chosen as the state-of-the-art benchmark since it is a well-known method to run MPC in an online fashion. \\
% {\color{red} Why we study these performance characteristics should be added}

For each trajectory, the $\ell_{2}$-norm of the difference between the set point for the feedback controller and the actual vehicle position is used to validate the accuracy of the MPC feedforward together with the state feedback controller. For the drone and the truck-trailer AMR, respectively 25~936 and 28~741 points are evaluated.
% Only a limited set of environments is chosen since the statistical distributions of this positional error per environment have a similar profile%{\color{red} in terms of mean and variance?},
% , thus capturing all necessary information needed to report on the performance of the feedback controller.

\begin{figure*}
    \centering
    % \hspace{-3ex}
    \includegraphics[scale=1.0]{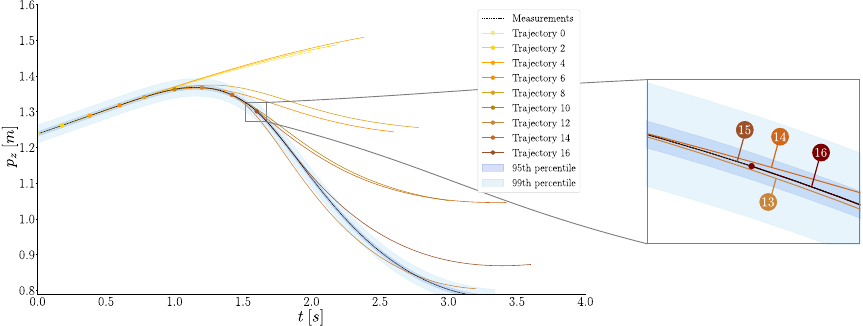}
    \caption{Illustration of the magnitude of (1) the feedback error compared to the actual trajectory and (2) the stitching strategy presented in Section~\ref{sec:ASAP_MPC} for a dataset of the quadrotor experiments. The 95th and 99th percentile represented on this figure are respectively 0.64 cm and 2.47 cm (specific to that dataset).}
    \label{fig:fb_error}
\end{figure*}

% Truck-trailer lab setup as in [ref IFAC]\\
% DodgeDrone simulation as in [ref IFAC]

% + to show that we can do it onboard:\\
% Truck-trailer all onboard with Jetson TX2\\
% Dodgedrone HITL with Jetson TX2

% \begin{flushleft}
% \begin{minipage}[b]{0.5\textwidth}
% \end{minipage}
% \begin{minipage}[b]{0.5\textwidth}
% [width=4.15cm,height=2.3cm]
% \end{minipage}
% \end{flushleft}

\subsection{Comparison with Real-Time Iteration Scheme}
In Section~\ref{section:related_work}, the RTI scheme is introduced as a feasible methodology for online optimal control in practical applications. Since it classifies as a FUR-MPC methodology, it is chosen as a benchmark for the presented ASAP-MPC framework. However, its implementation for the truck-trailer AMR manoeuvring proved to be unsuccessful, which can be attributed to the high level of non-convexity in the OCP and the large active set changes between subsequent solutions.
% \begin{itemize}
%     \item High level of \textbf{non-convexity} of the truck-trailer AMR OCP: difficult convergence and hence, high computation times for each QP iteration.
%     \item Significant changes in environment and thus \textbf{large active set changes} between subsequent solutions.
    % The high level of non-convexity of the optimisation problem underlying this problem originating from the complex vehicle kinematics and path constraints, prevents fast convergence of a single QP iteration.
    % \item The long computation time for a single QP iteration, makes it too slow for real time use. Additionally, it allows the environment to change more in between subsequent computations, making subsequent problems differ even more which makes it harder to converge fast enough.
    % \item The reference trajectory, as defined with corridors, shows significant jumps, especially when switching from one pair of corridors to another pair of corridors as the vehicle evolves along its trajectory.
% \end{itemize}

The non-convexity of the underlying OCP originates from the complex truck-trailer AMR kinematics. Additionally, when switching from one pair of corridors into the next, a significant change in active set occurs which violates the assumption that the SQP algorithm is warm-started close to the optimal solution and hence deteriorates the performance of the RTI scheme. Both complications lead to high computation times per SQP-iteration, diverging subsequent solutions from each other and thus limiting the warm-starting possibilities. The previous arguments covering non-convexity, active set changes and significant computation time also hold for the quadrotor application. The quadrotor dynamics reported in Table~\ref{tab:models} are inherently non-convex, a large active set change occurs between subsequent updates since there are previously unseen obstacles and as reported in~\cite{drone2023}, the computation times are even higher compared to the truck-trailer application~\cite{tt2023}. Therefore, implementing RTI for the drone application is expected to lead to similar results as for the truck-trailer application.
% \red{Since the drone model, detailed in Table~\ref{tab:models}, is more complex than the truck-trailer model, for which RTI does not even reach a sufficient performance level, the RTI scheme is expected to behave worse and thus considered as not applicable to the drone use case.}\\

% \red{RTI's inherent reliance on iterates converging to feasibility and optimality over multiple time steps might rapidly lead to intermediate solutions strongly violating corridor and dynamic constraints. These violations are expected to occur if a significant active set change arises. 
In contrast, ASAP-MPC, which allows for longer computation times and uses a fully converged solution each update instance, does not suffer from these problems and solves the problem meticulously, as shown in~\cite{tt2023}. However, to be able to actually compare both methods, a simplified case is studied for the manoeuvring of a truck without trailer, in a sequence of corridors without backwards manoeuvring. Additionally, to adhere to RTI's warm-starting assumption, the multi-stage approach from~\cite{tt2023} is adapted to a single-stage approach to reduce the active set changes while switching corridors. The main differences are a switch to a fixed time problem, with a horizon of six seconds, and the following change in corridor path constraints.

In the original formulation of~\cite{tt2023}, a fixed part of the control points in the horizon, i.e. a stage, is assigned to a specific corridor. In this new approach, each control point is individually assigned to a specific corridor and as such, can be individually reassigned to the next corridor as the horizon recedes. Consequently, the trajectory evolves more gradually into the next corridor. This avoids large active set changes, and thus large computation times, at corridor switches. The RTI scheme is implemented as an SQP algorithm with qpOASES as QP solver, initialising each QP problem with the previous solution and interrupting the SQP method after solving one QP problem. \\

% Originally, the consecutive parts of the trajectory were always assigned to a specific corridor. In this new approach, all the discrete vehicle poses along the trajectory get their own instance of either the first or the second corridor, while a high level planner makes sure which one to choose. This allows to initialize the vehicle using only the first corridor and limits the corridor switches to a maximum of one between each iterate. A high level planner checks after every iteration if any of the discrete vehicle poses is already in a second corridor, in which case the constraint related to this point is switched from the first corridor to the next one by updating the constraint parameters for that pose. Additionally, as the problem is no longer free time, the objective changes from minimal motion time to a quadratic penalization of the endpoint of the trajectory towards a local destination within the current corridor. 

\begin{figure}
    \begin{subfigure}[t]{0.24\textwidth}
        \centering
        \includegraphics[width=4.9cm]{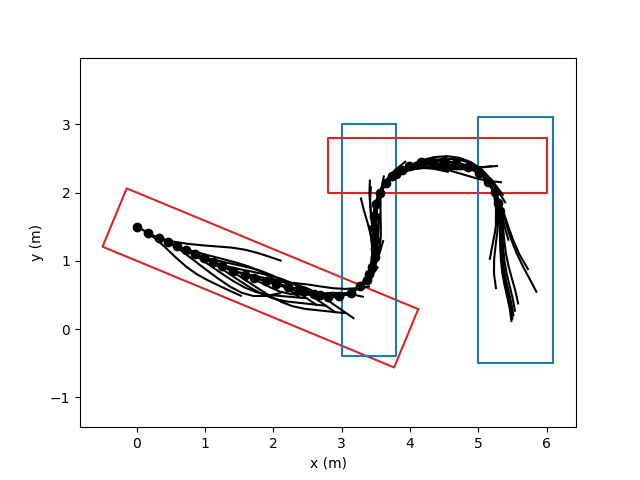} %width=\linewidth
        \caption{Subsequent RTI solutions}
        \label{fig:tt_RTI}
    \end{subfigure}
    \begin{subfigure}[t]{0.24\textwidth}
        \centering
        \includegraphics[width=4.9cm]{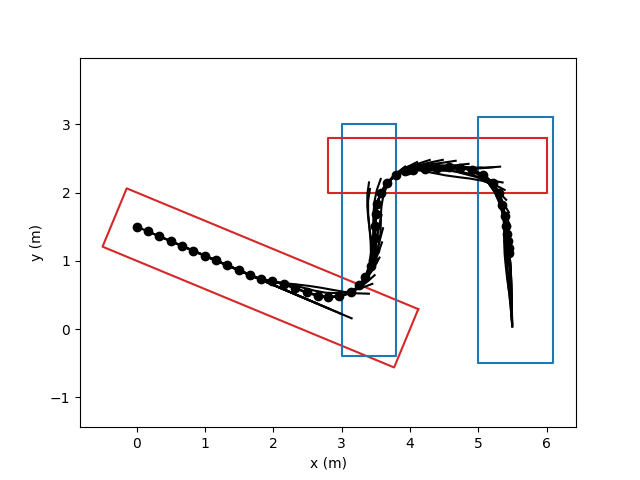} %width=0.47\linewidth
        \caption{Subsequent ASAP-MPC solutions}
        \label{fig:tt_ipopt}
    \end{subfigure}
    \caption{Comparison of all subsequent trajectories for RTI (stopping after one iteration) and ASAP-MPC (with a fully converged solution).}
    \label{fig:tt_RTI_ipopt}
\end{figure}

Both the computation times and the trajectories are compared for both ASAP-MPC (with Ipopt) and RTI for the single-stage approach. A remarkable difference can be found in computation time for both methods. The average computation time for Ipopt, taken over 50 subsequent time steps, is 8~ms, whereas the average computation time for qpOASES is 18~ms. As QP solvers typically strongly benefit from a proper initialisation, often using a previous solution, it is expected that solving the QP subproblem would be faster. However, as the computation time for a single iteration takes a reasonable amount of time, the truck has already progressed significantly. Hence, two subsequent problems differ more significantly than expected and therefore, finding a solution to the QP subproblem becomes more difficult which further deteriorates convergence of the QP subproblem in a subsequent step. Trajectories from both methods seem feasible with respect to the corridors, but it should be noted that the trajectories from qpOASES are not a solution to the original OCP and thus inhibit a risk that their solution might be infeasible with respect to the kinematic model or other constraints. On the other hand, Ipopt always converges to locally optimal trajectories for the original OCP. Furthermore, Figure~\ref{fig:tt_RTI_ipopt} clearly shows RTI's dependency on the trajectories converging to an optimum for the original OCP over multiple updates.

\section{Results and Discussion}
\label{section:results}
To compare ASAP-MPC with FUR-MPC and LUR-MPC, the update rate of FUR-MPC in both applications is chosen at 50 Hz, which is the rate necessary to stabilise and control the drone and the truck-trailer AMR. The update rate of LUR-MPC is taken equal to the maximal computation time in all the experiments for each of the applications separately. The very first computation time of each experiment is left out, as one could argue that this is an outlier without warm-starting of the MPC, and the first solution time is less critical since it could be computed offline.

\subsection{Drone in a Cluttered Environment}
\label{section:drone_results}
% Wat moeten we aanduiden op foto:
% - Jetson - Linux 18.04 wiht ROS Melodic
% - Ethernet kabel - connectie to PC
% - Koeling vinnen
% - Power Cable - 12V DC power ??

% Experiments:
% - 5 easy envs
% - 5 mediums envs
% - 2 runs per env

Table~\ref{tab:perc_fb_error} and Figure~\ref{fig:fb_error} show the 95th and 99th percentile of the drone position tracking error, both three-dimensional and for each direction separately. Compared to the dimensions of the vehicle and the environment ($r_{\text{drone}} = $ 30~cm, $r_\text{obst} \in [$10~cm, 300~cm$]$), the positional error at 99$\%$ of the points is deemed acceptable. Consequently, it is concluded that the assumption mentioned in Section~\ref{sec:ASAP_MPC} of taking a future on-trajectory state as the best prediction of the robot's state at that future moment is valid. To guarantee constraint satisfaction on deployment, the error can be estimated offline and taken into account in the constraint formulation as a conservative margin, e.g. in the collision avoidance constraint. The agile flight application allows for this margin to exist since it is visually noted that the free space between obstacles is often an order of magnitude larger than the 99th percentile.

% {\color{red} Add histogram to show that often we are close to the actual predicted position ?}
% {\color{red} Do we report some relative measure compared to the scale of the actual setpoint /velocity ? Relative measure in terms of time needed to correct it and thus how far off we are from the point ?}

% % z-direction
% 95th percentile: 0.2924761141184717 cm
% 99th percentile: 0.9829196650534835 cm

% % y-direction
% 95th percentile: 2.973267640918493 cm
% 99th percentile: 9.659668944776051 cm

% % x-direction
% 95th percentile: 9.885772764682766 cm
% 99th percentile: 14.18348112702367 cm

% % 3D
% 95th percentile: 14.35863389370021 cm
% 99th percentile: 19.60017187480119 cm

\begin{table}
\centering
\caption{95th and 99th percentile of the feedback error for the drone case.}
\label{tab:perc_fb_error}
    \begin{tabular}{lccccc}
        \toprule
        \textbf{Application} & \textbf{Percentile} & $ \| \Delta \vec{p} \|_{2} $ & $ \| \Delta p_{x} \|_{2} $ & $ \| \Delta p_{y} \|_{2} $ & $ \| \Delta p_{z} \|_{2} $ \\
        \midrule
        Drone & 95\textsuperscript{th} & 14.46\:cm & 10.01\:cm & 3.02\:cm & 0.306\:cm\\
         & 99\textsuperscript{th} & 20.03\:cm & 14.44\:cm & 10.07\:cm & 1.03\:cm\\
        \midrule
        Truck-trailer AMR & 95\textsuperscript{th} & 4.37\:cm & 2.52\:cm & 3.80\:cm & NA\\
         & 99\textsuperscript{th} & 5.77\:cm & 3.76\:cm & 4.78\:cm & NA\\

        \bottomrule
    \end{tabular}
\end{table}

% \red{Check if number is squared norm or normal norm}

Figure~\ref{fig:asap_onboard_drone} shows the computation times using Fatrop. A median update time of 45 ms is achieved, with a maximal deviation towards 542 ms. Comparing the results to the required rate by FUR-MPC, equal to 50 Hz, it clearly shows that in most cases this rate is not achieved and can thus not be guaranteed. Nonetheless, the fixed rate of LUR-MPC, chosen based on the maximal wall time of 542 ms, is too conservative and does not allow rapid responses to a dynamically changing environment. The ASAP-MPC framework makes it feasible to deal with this variation in update rate while still applying control inputs to the robot at the rate required to run on-line MPC-based trajectory generation onboard.

\begin{figure}
    \centering
    \includegraphics[width=1.0\linewidth]{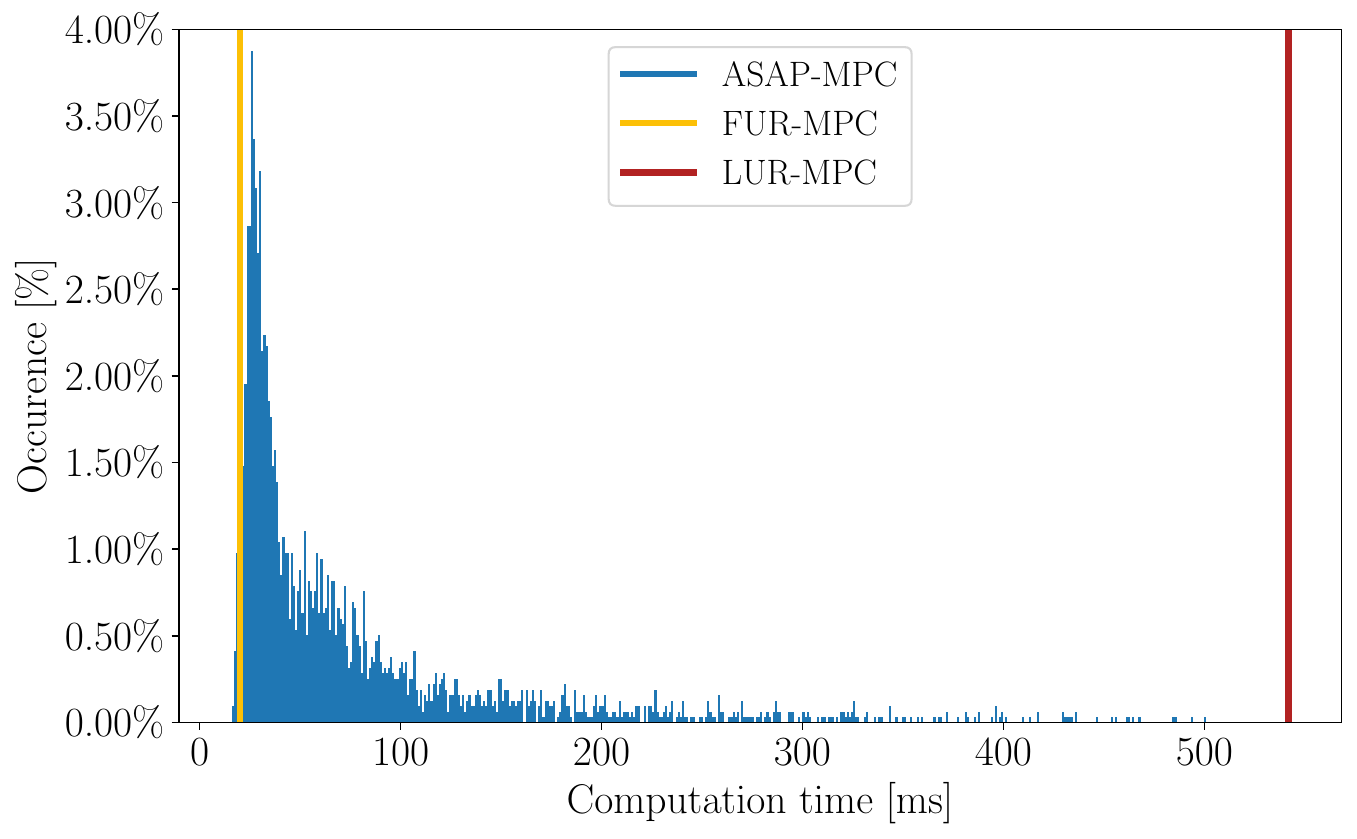}
    \caption{Online ASAP-MPC computation times for the drone application, with the required sampling times for FUR-MPC (20 ms) and LUR-MPC (542 ms), benchmarked on the Nvidia Jetson TX2.} % Excluding the first computation of a run.
    \label{fig:asap_onboard_drone}
\end{figure}

\begin{figure}
    \centering
    \includegraphics[width=1.0\linewidth]{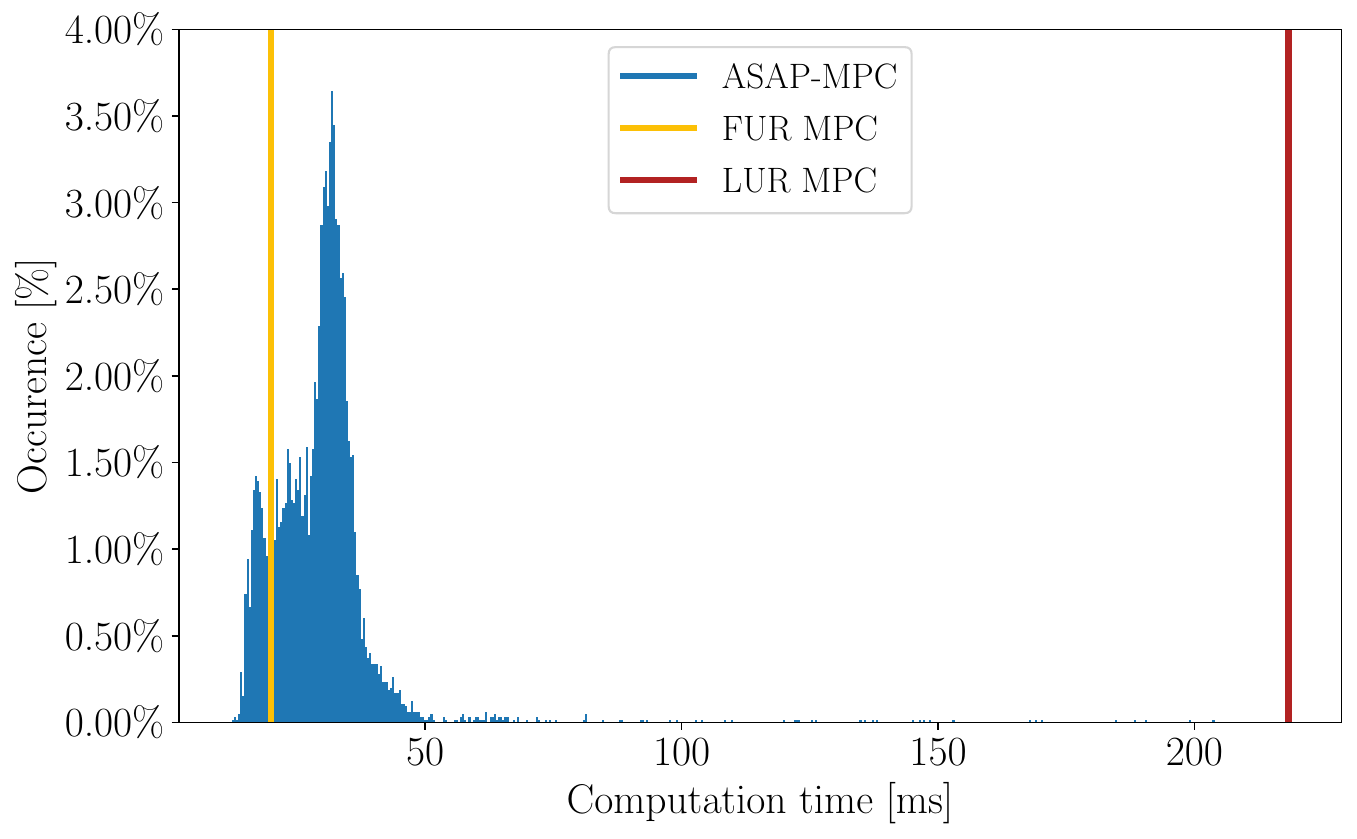}
    \caption{Online ASAP-MPC computation times for the truck-trailer AMR application, with the required sampling times for FUR-MPC (20 ms) and LUR-MPC (218 ms), benchmarked on the Nvidia Jetson TX2.} % Excluding the first computation of a run.
    \label{fig:asap_onboard_drone}
\end{figure}

\subsection{Truck-Trailer AMR Parking Manoeuvre}
Table~\ref{tab:perc_fb_error} also shows the 95th and 99th percentile of the position tracking error for the rear end of the trailer. Given the limited tracking error, adding a small margin as proposed in~\cite{tt2023}, suffices to conclude that the parking manoeuvre is successful without exceeding the road and parking space markings. Therefore, the positional error at 99$\%$ of the points is deemed acceptable. Again, it is concluded that the assumption mentioned in Section~\ref{sec:ASAP_MPC} of taking a future on-trajectory state as the best prediction of the robot's state at that future moment is valid.
% !TeX root = main.tex
\section{Conclusion}
\label{section:conclusion}

This work presented a Nonlinear Model Predictive Control scheme for motion planning, termed ASAP-MPC, capable of dealing with the computational delay that arises from solving an OCP online while still guaranteeing feasibility of the resulting solutions with respect to robot dynamics and various constraints. The ASAP-MPC methodology decouples high-rate trajectory tracking with low-rate trajectory generation. ASAP-MPC seamlessly combines them using smooth transitions between subsequent locally optimal trajectories, providing both disturbance rejection and swift reactions to changing environmental information. Consequently, it eliminates unwanted behaviour due to the computational delay, such as drifting because of plant-model mismatch and the aforementioned trajectory jumping phenomenon. These features come at the expense of requiring an additional feedback controller and having slightly higher update times of the reference trajectories compared to classical FUR-MPC approaches. \\

The ASAP-MPC algorithm is successfully implemented onboard on two complex, real-life applications: a trailer with truck manoeuvring in confined spaces and a drone agilely flying in a locally known, cluttered environment. On a simplified version of the truck-trailer application, RTI is implemented as a state-of-the-art online MPC approach. The presented framework's performance with respect to RTI is deemed better in terms of computation times, similarity of subsequent solutions and feasibility for this simplified example. Applying RTI to the more complex examples proved to be unsuccessful. \footnote{Software involved in creating the figures, the reported data and the implementations of the two discussed applications can be found on \url{https://gitlab.kuleuven.be/meco-publications/asap-mpc} in the \textit{as\textunderscore submitted\textunderscore tase} branch.}\\

The scheme's characteristics and assumptions surrounding smooth on-trajectory connections and trajectory update times are validated on both applications. The distribution of the feedback error and the trajectory computation times are studied over a significant amount of trajectories and the results experimentally show that the update points are reached with an acceptable accuracy and that the computational delay is both present and successfully dealt with using the ASAP-MPC scheme. \\

Towards future research, firstly a more robust method that has guarantees for finding a solution to the OCP can be found to further improve robustness of the framework. Secondly, automatic hyperparameter tuning of the MPC problem and the underlying solver, such as control horizon, through reinforcement learning is expected to improve both the planning times and the quality of the resulting trajectories. Lastly, the current feedback controller can be extended towards an MPC-tracking controller, e.g. using RTI, which is expected to improve the tracking error. This might be necessary in applications requiring tighter tracking requirements such as automatic welding and laser contouring.

% {\color{red} Future work: robuster method of motion planning that is guaranteed to give a solution, automatic hyperparameter tuning through reinforcement learning, improve implementation to make it easier-to-use, extend simple feedback controller to MPC or RTI-based tracking controller}

% Advantages:
% - Allow Optimal Control in MPC on complex examples, guaranteeing feasibility
% - Decoupling of high-rate trajectory tracking control with low-rate trajectory generation

% Disadvantages:
% - Additional linear feedback control
% - Slightly slower reactions to environment

% What has to be in it:
% - Prevent unwanted behaviour due to delay: drifting, trajectory jumping
% --> solutions: disturbance rejection, smooth transition between subsequent trajectories

% ===========================================

\section*{Acknowledgments}
This research was partially supported by Flanders Make, the strategic research centre for the manufacturing industry. VLAIO (Flanders Innovation \& Entrepreneurship Agency) is also acknowledged for its support. This work benefits from the ICON project FROGS: Flexible and Robust Robotic Gripping Solutions, the SBO project ARENA: Agile and Reliable Navigation , the SBO project FLEXMOSYS: Flexible Multi-Domain Design for Mechatronic Systems and the SBO project DIRAC: Deterministic and Inexpensive Realizations of Advanced Control. The authors also want to thank Alejandro Astudillo Vigoya for his extensive feedback to improve the written part of this work.

% ARENA, FROGS, DIRAC, FLEXMOSYS

% {\appendix[Proof of the Zonklar Equations]
% Use $\backslash${\tt{appendix}} if you have a single appendix:
% Do not use $\backslash${\tt{section}} anymore after $\backslash${\tt{appendix}}, only $\backslash${\tt{section*}}.
% If you have multiple appendixes use $\backslash${\tt{appendices}} then use $\backslash${\tt{section}} to start each appendix.
% You must declare a $\backslash${\tt{section}} before using any $\backslash${\tt{subsection}} or using $\backslash${\tt{label}} ($\backslash${\tt{appendices}} by itself
%  starts a section numbered zero.)}



\bibliographystyle{IEEEtran}
\bibliography{references}

% \end{thebibliography}

% \newpage

 

% \vspace{-33pt}
\begin{IEEEbiography}[{\includegraphics[width=1in,height=1.25in,clip,keepaspectratio]{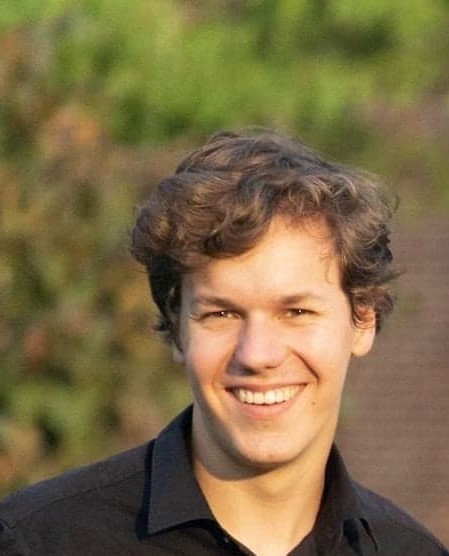}}]{Dries Dirckx} received the M.S degree in Mechanical Engineering at KU Leuven, Leuven, Belgium in 2019. In 2020, he received a Postgraduate Program degree in Biomedical Engineering at KU Leuven, Leuven, Belgium where he currently is pursuing a Ph.D. degree with the Department of Mechanical Engineering at the Motion Estimation, Control and Optimisation Research Team. His research interests include trajectory planning for robotic manipulators and quadrotor drones, diffusion models and geometric collision detection algorithms.
\end{IEEEbiography}
\vspace{-33pt}
\begin{IEEEbiography}[{\includegraphics[width=1in,height=1.25in,clip,keepaspectratio]{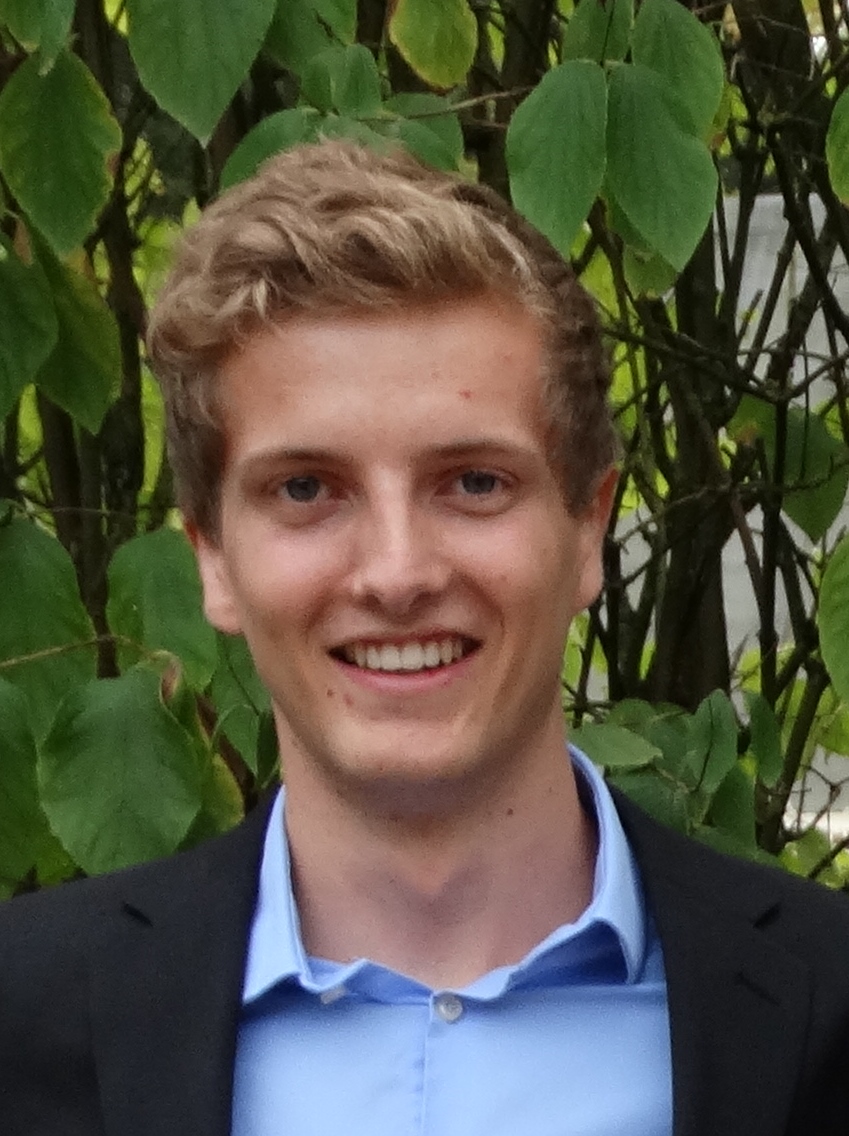}}]{Mathias Bos} is currently working towards a Ph.D. at the Motion Estimation, Control and Optimisation Research Team at KU Leuven, Belgium, where he also obtained his M.S. degree in Mechanical Engineering in 2019. His research covers the motion planning and control of autonomous vehicles, such as quadrotor drones, autonomous mobile robots, and self-driving cars.
\end{IEEEbiography}
\vspace{-33pt}
\begin{IEEEbiography}[{\includegraphics[width=1in,height=1.25in,clip,keepaspectratio]{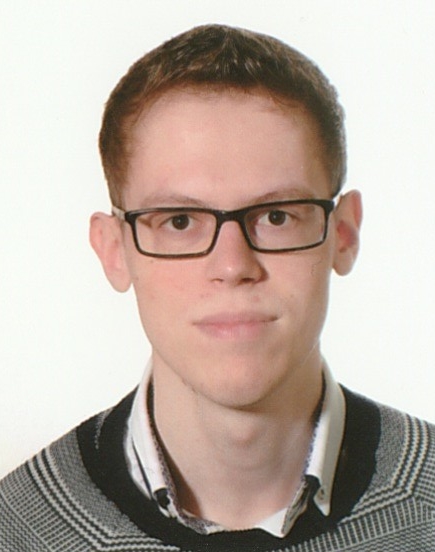}}]{Bastiaan Vandewal}
received the M.S. degree in Mechanical Engineering from KU Leuven, Belgium in 2017. In 2024, he received his Ph.D. degree at the Department of Mechanical Engineering, KU Leuven, as a member of the Motion Estimation, Control and Optimisation (MECO) Research Team. His research interests include the areas of control theory and motion planning with a focus on applications for autonomous vehicles.
\end{IEEEbiography}
\vspace{-33pt}
\begin{IEEEbiography}[{\includegraphics[width=1in,height=1.25in,clip,keepaspectratio]{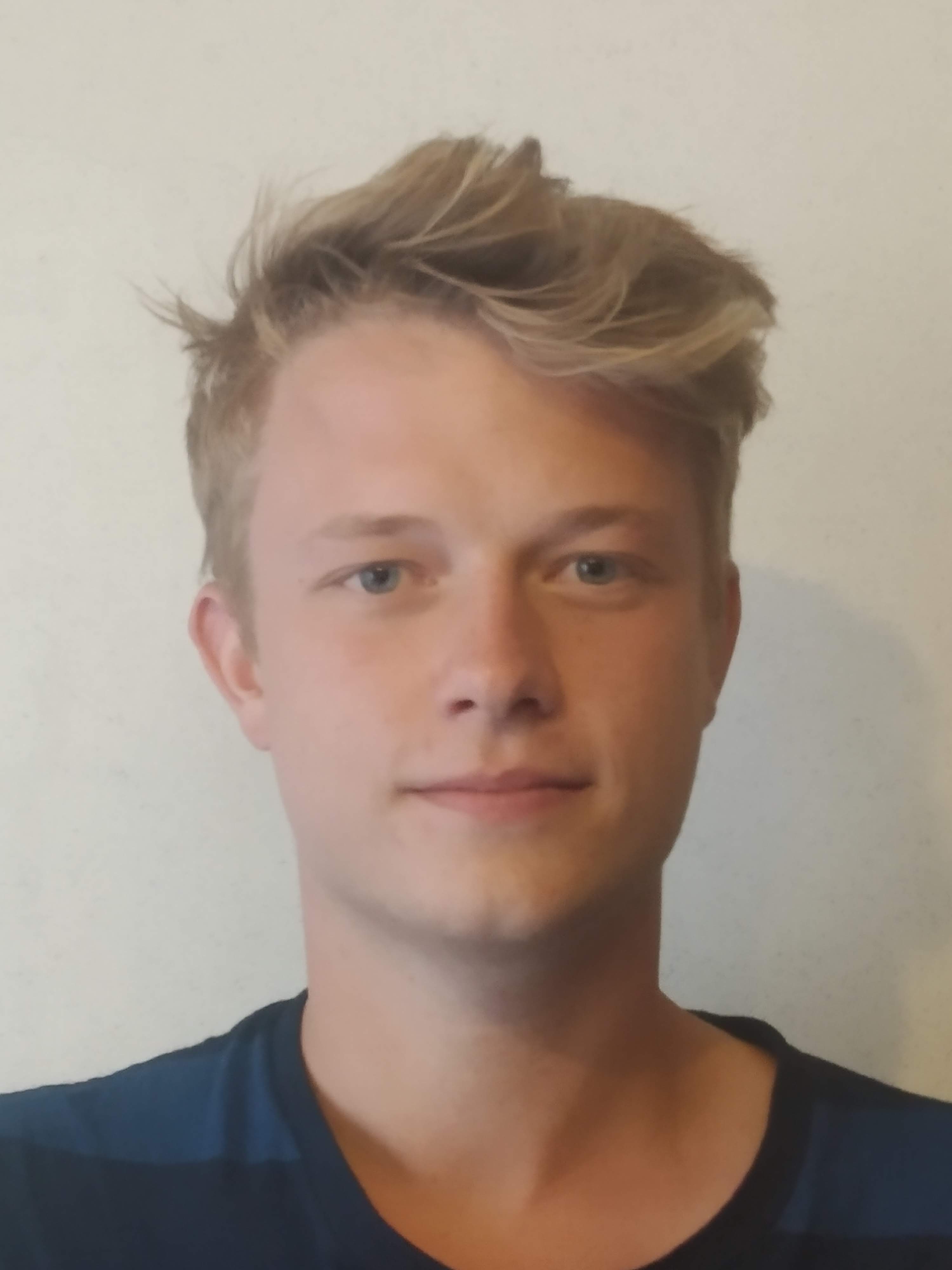}}]{Lander Vanroye}
obtained the M.S. degree in Mathematical Engineering, KU Leuven 2019. He is now a Ph.D. researcher at the Mechanical Engineering Department of the same university. His research is focused on the development of a numerical solver for optimal control problems.
\end{IEEEbiography}
\vspace{-33pt}
\begin{IEEEbiography}[{\includegraphics[width=1in,height=1.25in,clip,keepaspectratio]{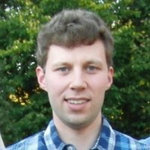}}]{Wilm Decr\'{e}}
received the M.S. and Ph.D. degrees in Mechanical Engineering from KU Leuven, Belgium in 2006 and 2011, respectively. He is a research manager at the Department of Mechanical Engineering of KU Leuven. His research interests include sensor- and optimisation-based control of robot systems, numerical optimisation algorithms and applications, learning and optimal control and estimation, and real-time and embedded software design.
\end{IEEEbiography}
\vspace{-33pt}
\begin{IEEEbiography}[{\includegraphics[width=1in,height=1.25in,clip,keepaspectratio]{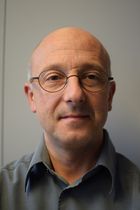}}]{Jan Swevers}
received the M.S. degree in Electrical Engineering and the Ph.D. degree in Mechanical Engineering at KU Leuven, Belgium, in 1986 and 1992, respectively. He is full professor in the Department of Mechanical Engineering of KU Leuven, and coordinates the Motion Estimation, Control and Optimisation Research Team. His research focuses on motion control and optimisation of mechatronic systems: optimal control of linear multivariate systems, iterative learning control, system identification,  optimal motion planning and embedded optimisation for motion control systems. He is a member of the MPRO core lab of Flanders Make@KU Leuven, Belgium. 
\end{IEEEbiography}


% % \bf{If you will not include a photo:}\vspace{-33pt}
% \begin{IEEEbiographynophoto}{Bastiaan Vandewal}
% \begin{IEEEbiographynophoto}{Mathias Bos}

% \vfill

\end{document}